\documentclass[10pt,twocolumn,letterpaper]{article}

\usepackage{Styles/iccv}     

\usepackage{graphicx}
\usepackage{enumitem}
\usepackage{tikz}
\usepackage{comment}
\usepackage{amsmath,amssymb} 
\usepackage{amsthm}
\usepackage{colortbl}
\usepackage[accsupp]{axessibility} 
\usepackage{bbm}
\usepackage{dsfont}
\usepackage{bm}
\usepackage{booktabs}
\usepackage{tcolorbox}
\usepackage{algorithm2e}
\usepackage{subcaption}
\usepackage{float} 
\usepackage{tabularx} 

\usepackage{blindtext}
\usepackage[export]{adjustbox}
\usepackage{pifont}
\usepackage{mdframed,lipsum}

\usepackage{tikz}
\usepackage{pgfplots}
\definecolor{darkgreen}{RGB}{31,182,83}
\usepackage[toc,titletoc,page]{appendix}
\usepackage[pagebackref=true,breaklinks=true,colorlinks,bookmarks=false]{hyperref}

\newcounter{problem}
\newenvironment{problem}%
{%
    \refstepcounter{problem}%
}{}

\usepackage[capitalize]{cleveref}
\crefname{section}{Sec.}{Secs.}
\Crefname{section}{Section}{Sections}
\Crefname{table}{Table}{Tables}
\crefname{table}{Tab.}{Tabs.}
\definecolor{LightGray}{gray}{0.9}

\newcommand\tabcaption{\def\@captype{table}\caption}
\newcommand\figcaption{\def\@captype{figure}\caption}
\makeatother
\definecolor{LightGray}{gray}{0.96}
 \iccvfinalcopy 

\ificcvfinal\fi
\pgfplotsset{compat=1.17}
\begin{document}

\title{Removing supervision in semantic segmentation \\with local-global matching and area balancing}

\author{Simone Rossetti$^{1,2}$ \qquad Nico Samà$^1$ \qquad Fiora Pirri$^{1,2}$\\
$^1$DeepPlants \qquad  {\tt\small @deepplants.com}\\
$^2$DIAG, Sapienza \qquad {\tt\small @diag.uniroma1.it} \\ 
}
\maketitle
\ificcvfinal\thispagestyle{empty}\fi

\begin{abstract}
Removing supervision in semantic segmentation is still tricky. Current approaches can deal with common categorical patterns yet resort to multi-stage architectures. We design a novel end-to-end model leveraging local-global patch matching to predict categories, good localization, area and shape of objects for semantic segmentation. The local-global matching is, in turn, compelled by optimal transport plans fulfilling area constraints nearing a solution for exact
shape prediction. 
Our model attains state-of-the-art in Weakly Supervised Semantic Segmentation, only image-level labels, with 75\%  mIoU on PascalVOC2012 \textit{val} set and 46\% on MS-COCO2014 val set.
Dropping the image-level labels and clustering self-supervised learned features to yield pseudo-multi-level labels,  we obtain an unsupervised model for semantic segmentation. We also attain state-of-the-art on Unsupervised Semantic Segmentation with 43.6\%  mIoU on PascalVOC2012 \textit{val} set and 19.4\% on MS-COCO2014 \textit{val} set. Code is available at \url{https://github.com/deepplants/PC2M}.
\end{abstract}


\typeout{-------------- INTRO --------------}
\section{Introduction}\label{sec:intro}
 Semantic segmentation is the computational task of discovering object categories in an image, localising them, and predicting their shapes and dimensions. Recent research has successfully proved that weaker or no supervision can overtake demanding pixel supervision. 

Weakly Supervised Segmentation methods (WSSS)  \cite{xu2022multi,du2022weakly,chen2022class,xie2022clims, RossettiECCV-2022} replace pixel supervision with image-level supervision. The image categories prior define the space of objects. This prior knowledge, combined with the intrinsic ability of deep networks to focus on regions of interest, allows localising class-discriminative regions (CAM \cite{zhou2016learning}). Keeping  CAM  for proper localisation imposes a tremendous effort to expand and clean CAMs attention maps for improving mask-shapes, such as resorting to affinity net \cite{ahn2018learning,wang2020self,zhang2021complementary,xu2021leveraging}.

\begin{figure}[t]
\centering
\includegraphics[width=0.99\linewidth]{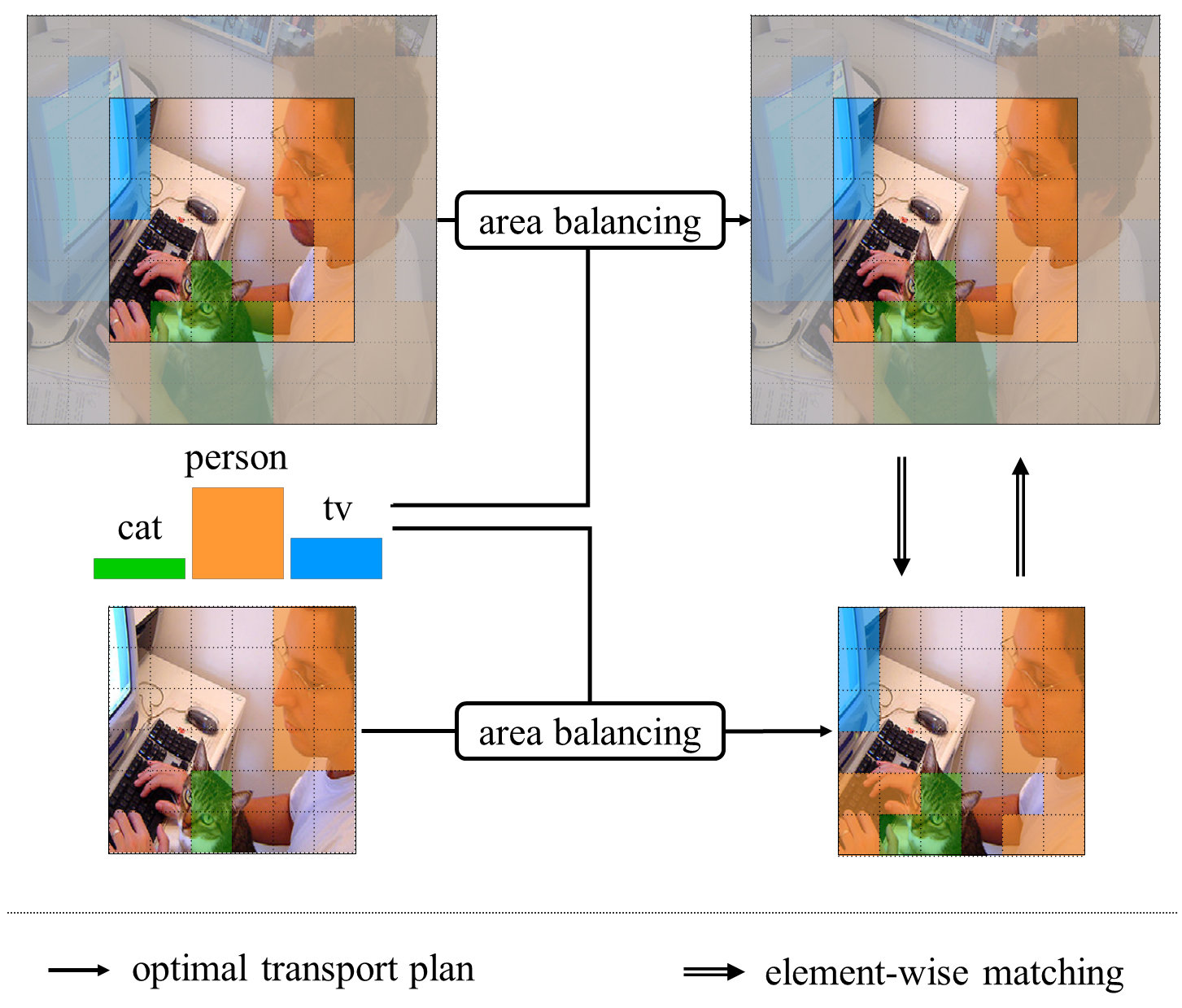}
\caption{Our method matches local and global predictions while satisfying estimated categories' area constraints. Left-side images are segmentation network predictions of the global (top) and local (bottom) aligned views of the image, which strongly disagree. An optimal transport plan updates the predictions (right-side images) to fulfil area constraints (bar chart). Element-wise matching drives the network towards local-global agreement. }
\label{fig:concept}
\end{figure}%
\begin{figure*}[h]
  \centering
       \includegraphics[width=10.5cm,height= 6.5cm]{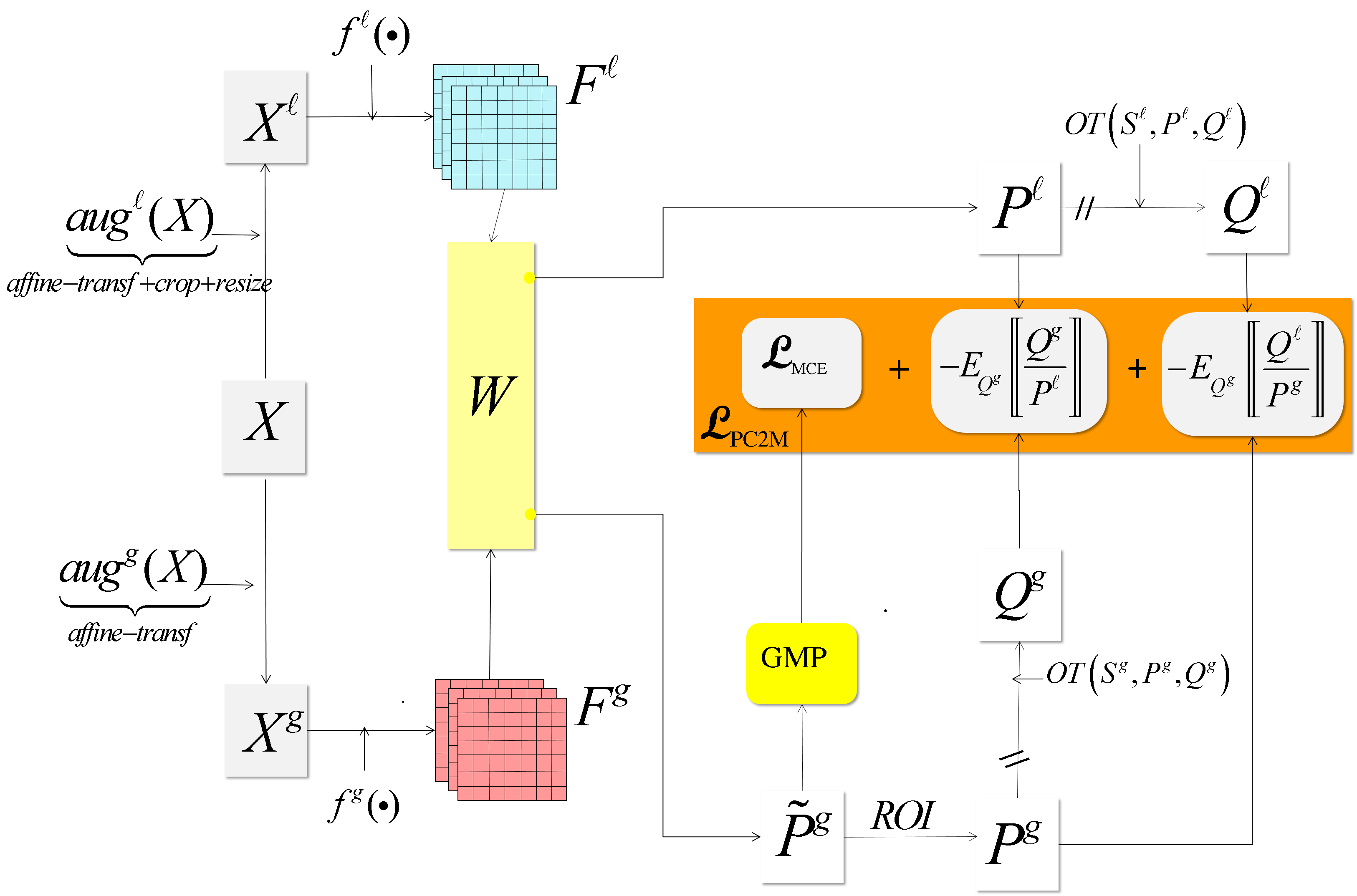}
     \includegraphics[width=6cm,height=6.5cm]{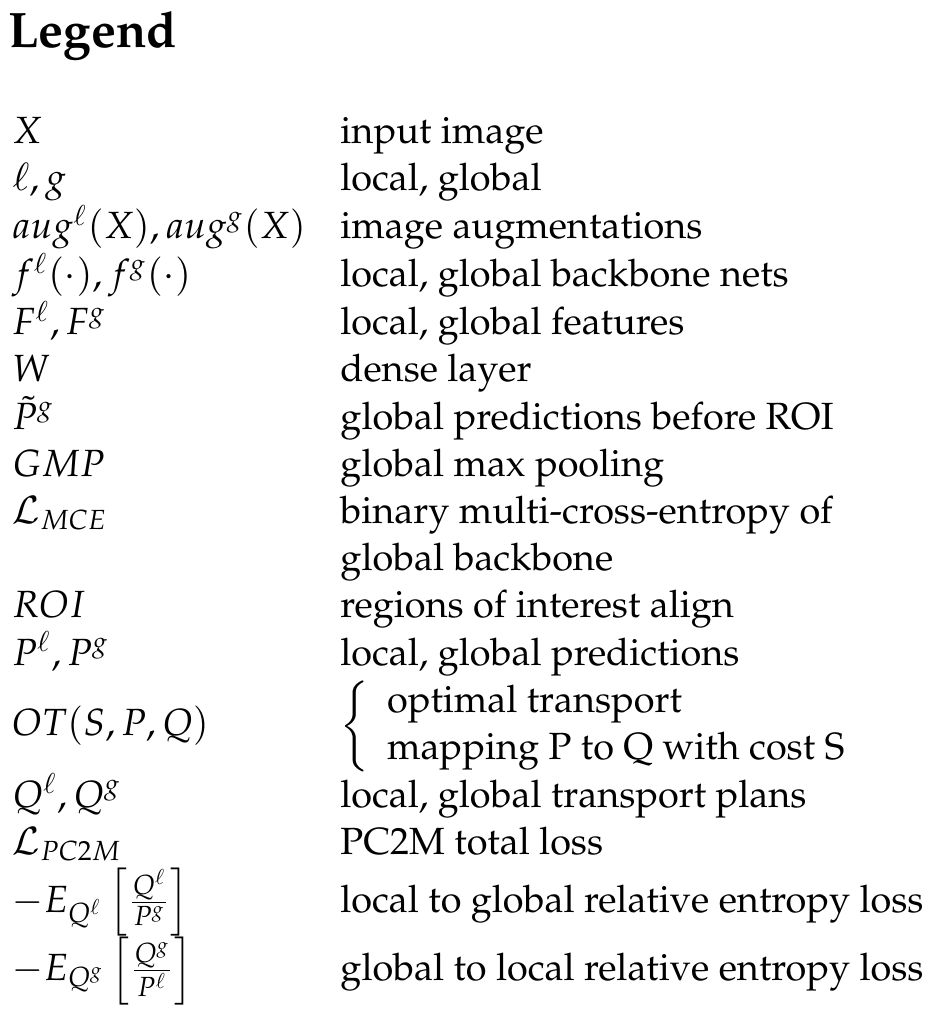}
    \caption{The proposed $\mbox{PC2M}$ model: A ViT-based backbone extracts the features from augmented views of the same image along two branches. The features are then projected to get the class probabilities for each patch in the views. The loss function is set up to minimise a trade-off between the image-level class prediction error and patch-level class prediction matching for each branch, where patch-level class assignments are estimated by solving an optimal transport problem.  }
    \label{fig:arch}
\end{figure*}

Some methods like AFA \cite{ru2022learning} and MCTFormer\cite{xu2022multi} have explored transformers \cite{dosovitskiy2021image}, focusing on the use of the class token for uplifting categories to patches, still committing to CAMs for localisation. So far, only \cite{RossettiECCV-2022} have shown that transformers can replace CAM using dense attention for localisation and patch category assignments. 
CAM ensures sound localisation, although not shape consistency, and to go beyond this bottleneck, these two semantic segmentation properties must be assessed together when supervision is weakened or removed.
We take appropriate steps in this direction, nearing the solution to sound localisation and mask shapes.
The concept of our work is shown in \Cref{fig:concept}. Here, we present Patch-Class to Masks (PC2M) lifting dimension, hence shape, close to supervised methods.

Unsupervised Semantic Segmentation (USS) methods \cite{ji2019invariant,ouali2020autoregressive,kim2020unsupervised,harb2021infoseg,cho2021picie,van2021unsupervised,melas2022deep,ziegler2022self,ke2022unsupervised,wenself} obtained so far successful results, following the self-supervised representation principles of pulling close similar data and pushing away different ones. Though here, as in dense self-supervised representations, data are not images but pixels, patches, or super-pixels (see, e.g. \cite{wang2021dense,li2021,pang2022unsupervised1,pang2022unsupervised2,xie2021propagate,wang2022exploring}). These methods leverage pixel similarity and spatial information encoding them into distinctive features corresponding to disjoint categories. Categories discovery relies upon clustering methods to group the learned features into the relevant dataset classes.

Unlike image classification, object detection and recognition, semantic segmentation displays the perception of object states with their semantics, such as their shape, position within the image vantage point, and relative dimension. Dense bottom-up methods such as IIC \cite{ji2019invariant}, MaskContrast \cite{van2021unsupervised}, Leopart \cite{ziegler2022self}, and SlotCon \cite{wenself} focusing on pixel similarities, though producing high-quality masks of object parts, struggle to group consistently discovered parts. 

To address this problem we  promote our WSSS method to USS, replacing image-level labels with pseudo-labels. The proposed model clusters features self-supervised on ImageNet \cite{russakovsky2015imagenet}
Our  PC2M, illustrated in Figure \ref{fig:arch},   advances the solution to the unsupervised mask-shape problem by handling the object area estimation for each category. 

The main contributions of this work are:

\noindent 
$\bullet$ We introduce PC2M an end-to-end network for semantic segmentation supervised with image-level labels. The improvements are due to: 

1) PC2M is formed by two branches, one developing on augmentations with image parts and the other with augmentations using most of the image, in so forcing the network to collate local with global information, which is beneficial for grouping object parts; 

2)  PC2M relative-entropy loss expands with terms contrasting the objects' area predictions from the two network branches with dense pseudo-labels estimated with transport plans.
The transport plans are computed by solving an Optimal Transport (OT) problem \cite{peyre2019computational}  with properly generated discrete metrics. 
An exponential moving average gently drives the network to predict the correct object areas; see the example in Figure \ref{fig:concept}.  
 
\noindent
$\bullet$ We establish new state-of-the-art performances in WSSS with 76.55\% and  75.7\% mIoU on PascalVOC2012 \cite{everingham2010pascal} train and test set (against the best supervised 89.0\% mIoU with DeepLabv3+Xception-65-JFT \cite{Chen2018DeepLabSI}), and 46.0\% on the MS-COCO2014 \cite{lin2014microsoft} val set. 

\noindent
$\bullet$ We introduce a principled top-down approach to USS. We show that replacing ground-truth labels with self-supervised pseudo-labels leads to a model attaining state-of-the-art performance in USS, with a mIoU of 43.6\% on PascalVOC-2012 and 19.42\% on MS-COCO2014 val sets. Our top-down method sheds light on the importance of category knowledge. We show that the absence or low precision of contextual information encoded in image-level labels is detrimental to our framework. We argue that improving unsupervised methods for high-quality label extraction is essential for building robust unsupervised segmentation models.


\typeout{-------------- State of The Art  --------------}
\section{Related Works}\label{sec:sota}
\begin{figure}
\centering
 \includegraphics[width=0.99\linewidth]{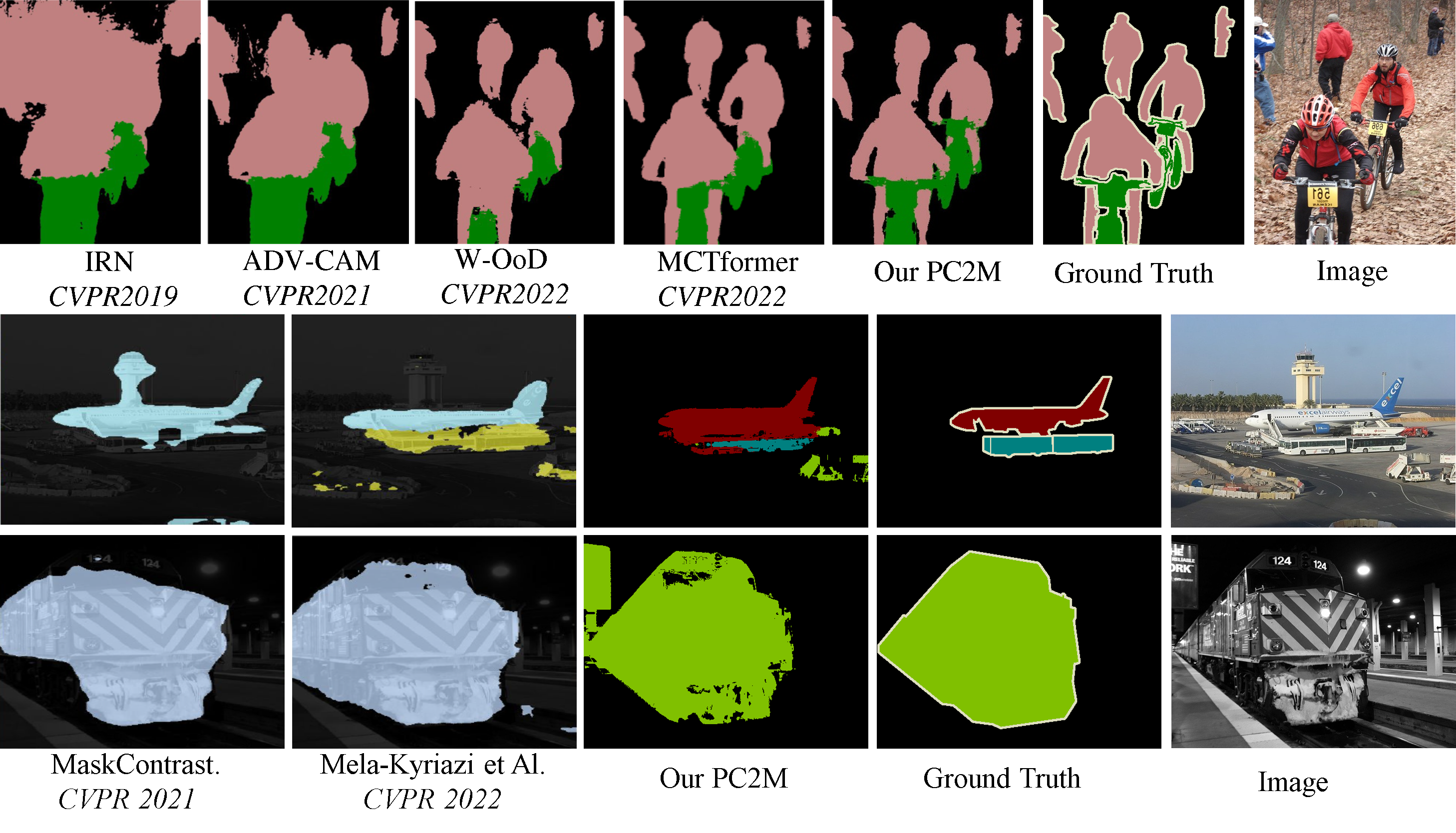}
\caption{Progress in removing supervision from Semantic Segmentation on the PascalVOC2012 dataset. The first stripe shows the progress of the WSSS since CVPR-2019. Ground-truth class labels are available here, and classes are consistently colour-coded because image-class labels are available. 
The two stripes below show the USS results.   Here, the class labels are not available. 
Class predictions from our method PC2M are correct in these images, and the shapes take advantage of our model learning object dimensions. The pseudo-masks of MaskContrast \cite{van2021unsupervised}, taken from \cite{melas2022deep}, do not use the PascalVOC colour code. }
\label{fig:prima-progress}
\end{figure}%

Figure \ref{fig:prima-progress} shows the progress of the WSSS and USS research streams in the most recent years. 

\noindent
{\bf Weakly supervised semantic segmentation.}
Recent methods of weakly-supervised segmentation predict pseudo-masks of the object categories with only image-level supervision. 
After \cite{zhou2016learning} introduced Class Activation Maps (CAM)  using a global average pooling layer in their CNN architecture, CAM has been massively used in WSSS \cite{chen2022class,lee2022weakly,kweon2021unlocking,stammes2021find,lee2021anti}. CAM-based methods often rely on mask-refinement procedures to enhance the accuracy of the baseline models, such as dense CRF \cite{krahenbuhl2011efficient}, and affinity maps \cite{ahn2018learning}. Pseudo-mask generation has been contaminated by self-supervised learning in \cite{wang2020self} via downstream tasks and transformations, ensuring CAM features equivariance. 
 Contrastive representation learning, for dense pixel embedding, is used in RCA \cite{zhou2022regional}, C2AM \cite{xie2022c2am}, and in PPC \cite{du2022weakly}. Lately, transformer-based architectures have gained traction as an alternative to CNN, such as  MTCformer \cite{xu2022multi}, AFA \cite{ru2022learning}, and ViT-PCM \cite{RossettiECCV-2022}  revealing an improved capacity of ViT in creating complete and well-localised masks. While both \cite{xu2022multi} and \cite{ru2022learning} use CAM to refine their pseudo-masks, ViT-PCM predicts categorical distributions at the patch level.

\noindent{\bf Unsupervised semantic segmentation.}
USS assigns a semantic label to each image pixel without human annotations, implying the need for multi-label support to discriminate several regions in the image. 
There are two lines of research on USS. The first is followed by MaskContrast \cite{van2021unsupervised}, focusing on dense pixel embedding using contrastive learning with initialised saliency maps as priors. MaskContrast is the first method to consider USS. 
The second line of research is grouping and clustering extended to pixels and patches. Leopart \cite{ziegler2022self} uses DINO features \cite{caron2021emerging} and learns an embedding for grouping similar pixels into object parts. Parts are then clustered according to the foreground and community detection. 
Clustering object parts incurs the risk of the same parts being assigned to the wrong objects. 
 PiCIE \cite{cho2021picie} incorporates  invariance to photometric and equivariance to geometric transformations. IIC \cite{ji2019invariant} uses mutual information for clustering and applies it to image patches. Similarly, \cite{ke2022unsupervised} start with classical grouping from  OWT-UCM \cite{arbelaez2009contours}, a grouping method that can hardly be leveraged for large-scale datasets. The method of \cite{melas2022deep} is the first one that elegantly solves the multi-label problem. Indeed, \cite{melas2022deep} extracts DINO-ViT-Base \cite{caron2021emerging} features to find the image segments by computing the gap between the non-null eigenvalues of the features Laplacian. In our top-down approach we follow a methodology similar to \cite{melas2022deep}, to  generate pseudo-labels supervising an end-to-end network  for USS.

\typeout{--------------MethodNewNew--------------}
\section{Our Approach}\label{sec:method}

This section presents our method by first discussing the model achieving state-of-the-art semantic segmentation with image-level label supervision. We further show how to expand our framework to unsupervised segmentation.

Similarly to \cite{RossettiECCV-2022}, which has shown excellent ability in localising class-relevant image segments given image-level supervision, we define a two-branch network based on ViT \cite{dosovitskiy2021image}.
By feeding different augmented views (one seizing most of the original image and one randomly cropping parts of it) to the two branches, we elicit representations for global and local views of a particular object category. The pooled predictions of the global branch are contrasted with the ground-truth image-level labels.
We use both branches' un-pooled patch predictions to yield dense pseudo-labels computed with the Sinkhorn-Knopp algorithm \cite{cuturi2013sinkhorn}. Namely, we formulate an Optimal Transport problem \cite{peyre2019computational} to constrain the dense pseudo-labels to assign each patch to a single class. Additionally, each patch assignment is balanced by the estimated normalised area of each class in the dataset. Since this area is unknown, we begin the training under the hypothesis that each object has an area proportional to the relative frequency of its category and slowly update it with the network empirical estimates.
Finally, we set up a contrastive loss between each branch patch prediction and the computed dense pseudo-labels.

\subsection{Preliminaries}

Let $D {=} \{X_1,\ldots, X_T\}$, with $X {\in} {\mathbb R}^{h\times w \times 3}$,  be a specified dataset with class labels $C {=}\{c_0,\ldots, c_k\}$,  with $|C|$ the cardinality of $C$. In the multi-label segmentation setting   $labels(X){\subseteq} C$. 
For each input image $X$, we create two views: one global view $X^g$ retaining most of the original image, and a local view $X^{\ell}$ obtained by randomly cropping and resizing $X$. The two views are augmented by randomly sampling a transformation from a set that includes jittering and affine transformations.  

We consider an encoder as the backbone network indicated by $f$; although their weights are shared (see Figure \ref{fig:arch}), we refer to the two branches as $f_g$ and $f_{\ell}$ for clarity.

Let $K{=}(n/d)^2$ with $n{=}w{=}h$ be the number of encoded patches of dimension $d$ for a specific augmented view $aug(X)$. Let $N {=} b{\cdot} K$ be the number of samples we consider in a batch of each branch $f^g, f^{\ell}$.   Let $F^{g}{\in} {\mathbb R}^{N\times e}$ and $F^{\ell}{\in} {\mathbb R}^{N\times e}$ be the two feature matrices of the augmented views conveyed from each branch. Each feature vector $F_i^t\in {\mathbb R}^e$,  $t {\in} \{g,\ell\}$, specifies the feature encoding a patch $u_i^t$ of a sampled augmented view from branch $f^t$.    The features $F^{t}, t{\in}\{g,\ell\}$, are projected on a dense layer with $|C|$ units shared by the two network branches $f_g$ and $f_{\ell}$. 
The estimated weights  are $W{\in}{\mathbb R}^{e\times |C|}$. 

$F_i^t W^{c_j}$ is the un-normalised cosine similarity between the features of the encoded $i$-th patch  and the $c_j$ class represented by the weight vector $W^{c_j}$,  the posterior probability  that the patch $u_i^t$  label $y^t_i$ is of class ${c_j}$ is:
\begin{equation}\label{eq:softmax}
p(y_i^t = c_j |u_i^t, W,F^t) = \frac{\exp (F_i^t W^{c_j})}{\sum_{c \in C}\exp(F_i^t W^{c})}
\end{equation} 
\noindent
which is the $(i,j)$-th element  of the prediction  matrix $P^t{=}\mathrm{softmax}(F^t W)$, $t{\in} \{g,\ell\}$. $P^t{\in} {\mathbb R}_{+}^{N{\times} |C|}$ enforces a categorical distribution for each patch $P^t_{i}$:  
$\sum_{j=1,\ldots, |C|}P_{i,j}^t {=} 1$. 

\vskip 1 \baselineskip
\noindent
{\bf Optimization objective} 
We first define the  match-loss, given $f_g$ and $f_{\ell}$, as follows:
\begin{equation}\label{eq:loss}
\begin{array}{lll}
     {\mathcal L}_{match}(P^g,P^{\ell}) &{=}& {-}{\mathbb E}_{Q^g}[\log P^{\ell}]  {-} {\mathbb E}_{Q^{\ell}}[\log P^{g}]{+}\\
 && H(Q^{\ell}) {+} H(Q^{g}) 
\end{array}
\end{equation}
Here, $Q^{\ell}$ and $Q^{g}$ are probability matrices with, respectively,  marginal distributions $(\frac{1}{N}{\mathds 1}_N, {\boldsymbol\alpha^{g}}){\in}{\mathbb R}_{+}^{N} {\times} {\mathbb R}_{+}^{|C|}$, and $(\frac{1}{N}{\mathds 1}_N,{\boldsymbol\alpha^{\ell}}){\in}{\mathbb R}_{+}^{N} {\times} {\mathbb R}_{+}^{|C|}$, generated according to conditions specified in the next section. A stop-gradient operation avoids the gradient back-propagation of the $H(Q^{\ell, g})$ entropy terms. The complete objective is: $$\mathcal{L}_{PC2M} {=}\mathcal{L}_{MCE} {+} \mathcal{L}_{match}$$
Here $\mathcal{L}_{MCE}$ is the standard multi-label binary cross-entropy loss, as in \cite{RossettiECCV-2022}.
\subsection{Optimal Transport for computing $Q^{g}$ and $Q^{\ell}$}
In this section, we describe how we generate the coupling matrices \cite{thorisson1995coupling}  $Q^g, Q^{\ell} \in {\mathbb R}_{+}^{N\times|C|}$ used in the match-loss, with Optimal Transport (OT). 

OT with entropic regularization \cite{peyre2019computational,cuturi2018semidual}  is defined as:
\begin{equation}\label{eq:OT}
\displaystyle{L^{\epsilon}_S({\bf c}, {\bf d}) = \min_{Q\in U({\bf c}, {\bf d})} \langle Q, S\rangle_F -\epsilon H(Q) }
\end{equation}

\noindent
Here $H(Q){=}{-}\sum_{i,j}Q_{i,j}(\log(Q_{i,j}){-}1)$  is the entropy of a coupling matrix,  $\epsilon{>}0$ is a weight of the entropy regularization, $\langle Q, S\rangle_F{=} tr(Q^{\top}S)$,  $S$ is a cost matrix, ${\bf c}$ and ${\bf d}$ are probability vectors in the probability simplices with $s$ and $n$ bins,  and the polytope
\begin{equation}\label{eq:couplings}
\displaystyle{U({\bf c}, {\bf d}) =\{Q\in {\mathbb R}_{+}^{n\times s} \ | Q {\mathds 1}_{s} ={\bf c} \mbox{ and } Q^{\top} {\mathds 1}_n ={\bf d}\}}
\end{equation}
\noindent
provides the mass conservation criterion \cite{peyre2019computational}. Solutions
of  the linear program in (\ref{eq:OT}) find a transportation plan $Q$ of minimal cost $S$ such that the marginal distributions of $Q$ are the discrete measures ${\bf c}$ and ${\bf d}$. In practice, (\ref{eq:OT})  ensures a unique solution of the OT problem converging to the maximal entropy transport coupling for a small regularisation $\epsilon$,  see \cite{peyre2019computational,cuturi2018semidual}.

Using the above results, and given ${\boldsymbol\delta}{=}\frac{1}{N}{\mathds 1}_N$, we define the two couples of probability vectors $({\boldsymbol\delta},{\boldsymbol\alpha^{g}})$ and $({\boldsymbol\delta},{\boldsymbol\alpha^{\ell}})$, generating the transportation plans $Q^g$ and $Q^{\ell}$. In the following, the computation of ${\boldsymbol\alpha^g}$ is the same as that of ${\boldsymbol\alpha^{\ell}}$ in the two branches; therefore, we drop the superscripts. 

Since we interpret $Q^{\ell, g}$ as the dense pseudo-labels for a specific algorithm iteration, the specified marginals ${\boldsymbol\alpha}$ over the class dimension can be deemed the vector of the normalised areas for each class in the full dataset. Given the image labels, we re-scale the class areas $\boldsymbol\alpha$ at each iteration to account for the class frequencies in the batch with respect to those in the dataset. Therefore, its exact form is\footnote{Each operation is intended element-wise}:  
\begin{equation}
    {\boldsymbol\alpha} =\frac{1}{Z}\frac{{\boldsymbol\nu_{b}}}{{\boldsymbol\nu_D}}\Tilde{{\boldsymbol\alpha}}
\end{equation}
Here $\Tilde{{\boldsymbol\alpha}}$ is the estimated class areas distribution, $\boldsymbol{\nu}_D$ and $\boldsymbol{\nu}_b$ are the relative class frequencies in the dataset and in the current batch. $Z$ is a normalization factor ensuring $\boldsymbol\alpha$ is a discrete measure.
We initialise  $\Tilde{\boldsymbol\alpha}$ uniformily as $\Tilde{\boldsymbol\alpha}_0{=}\boldsymbol\nu_D$. 

At each epoch $m{>}0$,  $\Tilde{\boldsymbol\alpha}$ is updated offline to the dataset's current predictions for each $X$ with an exponential moving average (EMA). Where the probability density vector of each $X$ with respect to class $c$ is:
\begin{equation}\label{eq:sumarea}
q_m^c(X){=}\displaystyle{\frac{1}{K}\sum_{k=1}^{K} p(y_k{=}c| u_k,F,W^c)}
\end{equation}\label{ema}
For all $X$ in the dataset $D$ the update of the $c$-th element of $\Tilde{\boldsymbol\alpha}$ at epoch $m$ is:
\begin{equation}\label{eq:area-update}
\Tilde{\alpha}_m^c =  (1{-}\gamma) \Tilde{\alpha}_{m{-}1}^c  {+} \gamma \frac{1}{T}\sum_{X{\in}D} q_{m-1}^c(X)
\end{equation}
Here $\gamma$ is the unit area momentum rate. For the optimization (\ref{eq:OT}) the solution proposed in \cite{cuturi2013sinkhorn,peyre2019computational}  finds the optimal $Q{\in} U({\boldsymbol\delta}, \boldsymbol\alpha)$ by computing two (unknown) scaling variables $({\bf u},{\bf v}){\in}{\mathbb R}_{+}^{N} {\times} {\mathbb R}_{+}^{|C|}$, 
such that $Q {=} \mbox{diag}({\bf u}){\mathcal K}\mbox{diag}({\bf v})$, with ${\mathcal K} {=} \exp(-\frac{S}{\epsilon})$, a Gibbs kernel, and $S$  a cost function. Here, to obtain  $Q^{g}$ and $Q^{\ell}$ we define the two kernels to be the predictions $P^{g}$ and $P^{\ell}$, that is, our cost matrices are, for $t{\in}\{g,\ell\}$: $S^t{=} {-}\log\left(P^t\right)$.
The scaling vectors for $Q^{g}$ and $Q^{\ell}$ are computed by Sinkhorn's algorithm presented in \cite{cuturi2013sinkhorn}.

\vskip 0.5\baselineskip
\noindent
 It is easy to see that if  $1/(T{\cdot} K)\sum_{j\in T\cdot K} P_j^{t,c} {=} \Tilde{\alpha}_m^c$ for all $c{\in}C$, with $K$ the number of patches in an image,  then the Sinkhorn algorithm stops updating, a condition we call convergence for our method. In fact, since ${\mathcal K} {=} (P^{t})^{1/\epsilon}$, with $\epsilon {\approx} 1$ we have, dropping the  superscript $t{\in}\{g,\ell\}$:
\begin{equation}\label{eq:converg1}
{\mbox{diag}({\mathds 1}_{K})} {[P^{1/\epsilon}]^{\top}}{\mbox{diag}({\mathds 1}_{TM})}({\mathds 1}_{TM})\propto \Tilde{\boldsymbol\alpha}_m
\end{equation} 
up to a constant.

On the other hand, the EMA stops updating when for all $c{\in} C$:
\begin{equation}\label{eq:converg2}
\Tilde{\alpha}_m^c = \Tilde{\alpha}_{m{-}1}^c = \frac{1}{T}\sum_{X{\in}D} q_{m-1}^c(X) = \frac{1}{T{\cdot} K}\sum_{j\in T\cdot K} P_j^{c}
\end{equation}
as can be derived from \Cref{eq:area-update}.
Hence, the network estimated marginal distributions over the class dimensions already match the prescribed distribution $\Tilde{\boldsymbol\alpha}_m$. Moreover, the marginal of the network predictions over the patches dimension matches $\boldsymbol\delta$ by construction\footnote{We further discuss these equalities and convergence in Appendix \ref{sec:dist_a}}. 

\subsection{Unsupervised Segmentation} 
We withdraw the image-level labels and compute the pseudo-multi-labels using self-supervised features to make the above method unsupervised. Similar to \cite{simeoni2021localizing} and \cite{melas2022deep} we use DINO \cite{caron2021emerging} self-supervised features. 

For each image in $D$, split in $n$ patches, we construct a graph with vertices $V{=}\{\phi(X_i^j)\}_{i=1, ..., T}^{j=1, ..., n}$ and edges $E{=}\{w(X_i^j, X_k^l)\}$ where $\phi$ is a self-supervised pre-trained feature extractor and $w$ is the pairwise affinity function:
\begin{equation}
    w(X_i^j, X_k^l) = \phi(X_i^j)^\top\cdot\phi(X_k^l)
\end{equation}
We compute the normalized Laplacian of the graph as $L=D_m^{-1/2}(D_m - A)D_m^{-1/2}$, where $D_m$ is the degree matrix and $A_{n(i-1)+j, n(k-1)+l} {=}w(X_i^j, X_k^l)$. After calculating the eigendecomposition of the graph Laplacian, the eigenvectors are clustered to obtain homogeneous regions, which are cropped and resized to a standard dimension. For each crop of $X_i$, we compute its CLS token feature and finally run $k$-means clustering on all the crops in $D$ with $k=|C|$. The cluster memberships of the crops define the desired pseudo-labels.

\typeout{----------------Experiments --------------------}
\section{Experiments}\label{sec:exper}
\typeout{----------------File Experiments --------------------}
We evaluate the effectiveness of our approach by performing experiments on the single semantic segmentation task but differentiating between two supervision strategies. First, we consider the weakly supervised case, and we further analyse the effectiveness of the unsupervised one.

\subsection{Datasets and Metrics}
For all experiments, we report the mIoU metric for the semantic segmentation task on PascalVOC2012 \cite{everingham2010pascal} (20 classes) augmented with the SBD dataset \cite{hariharan2011semantic}. We also evaluate our method on MS-COCO2014 \cite{lin2014microsoft} (80 classes).
We initialize our backbone as a Vision Transformer \cite{dosovitskiy2021image}, namely ViT-B/16, with weights pre-trained on ImageNet \cite{russakovsky2015imagenet}, following \cite{RossettiECCV-2022} for the WSSS task. For unsupervised segmentation, we start from a DINO \cite{caron2021emerging} initialization, similarly to \cite{ziegler2022self, melas2022deep}.

\subsection{Implementation Details}
\noindent

After one warm-up epoch, during which the initialized weights are frozen, we unfreeze the last five layers of the network and train the model on 2 NVIDIA RTX A6000 GPUs. For the PascalVOC2012 dataset, the model has been trained for 45 epochs, with a total training time of 4h 10min, while for MS-COCO2014, the training phase lasted 4h 50min and ran for 10 epochs. We discuss the algorithmic complexity of our method in Appendix \ref{sec:cost}. We use Adam \cite{kingma2014} optimizer and scale the initial learning rate $10^{-3}$ by $10^{-1}$ after the warm-up step.
At the end of every epoch, we update the target distributions for each class. The algorithm pseudo-code is presented in Appendix \ref{sec:alpha}.  As specified in Eq. \ref{eq:area-update}, we employ an exponential moving average (EMA). We modulate the EMA update via a momentum parameter $\gamma$ and discuss its effects in \Cref{sec:abl}.

\subsection{Ablation Studies}\label{sec:abl}
In this paragraph, we investigate the relative importance of the main components of our method by systematically ablating them. 

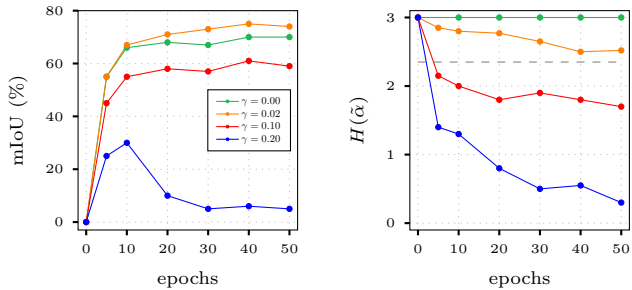
\begin{figure}[h!]
  \centering
  \begin{subfigure}[h]{0.99\textwidth}
    \begin{tikzpicture}
    \begin{axis}[
    title style={},
    xlabel={epochs},
    ylabel={mIoU (\%)},
    xmin=-2, xmax=52,
    ymin=-3, ymax=80,
    tick align=outside,
    xtick={0, 10, 20, 30, 40, 50},
    ytick={0, 20, 40, 60, 80},
    legend style={at={(0.97,0.5)},anchor=east},
    ymajorgrids=true,
    xmajorgrids=true,
    grid style=dotted,
    width=4.5cm,
    height=4.5cm,
    label style = {font=\scriptsize},
    legend style={nodes={scale=0.4, transform shape}},
    tick label style = {font=\tiny},
    every major tick/.append style={thick, major tick length=2pt, black},
    xtick pos=left,
    ytick pos=left,
    legend image post style={scale=0.5}
    ]
    \addplot[
        color=darkgreen,
        mark=*,
        mark options={solid},
        mark size=1pt
        ]
        coordinates {
        (0,0) (5,55) (10, 66) (20, 68) (30, 67) (40, 70) (50, 70)
        };
        \addlegendentry{$\gamma=0.00$}
    \addplot[
        color=orange,
        mark=*,
        mark options={solid},
        mark size=1pt
        ]
        coordinates {
        (0,0) (5,55) (10, 67) (20, 71) (30, 73) (40, 75) (50, 74)
        };
        \addlegendentry{$\gamma=0.02$}
    \addplot[
        color=red,
        mark=*,
        mark options={solid},
        mark size=1pt
        ]
        coordinates {
        (0,0) (5,45) (10, 55) (20, 58) (30, 57) (40, 61) (50, 59)
        };
        \addlegendentry{$\gamma=0.10$}
    \addplot[
        color=blue,
        mark=*,
        mark options={solid},
        mark size=1pt
        ]
        coordinates {
        (0,0) (5,25) (10, 30) (20, 10) (30, 5) (40, 6) (50, 5)
        };
        \addlegendentry{$\gamma=0.20$}
    \end{axis}
    \end{tikzpicture}
    \hspace{6pt}
    \begin{tikzpicture}
    \begin{axis}[
    title style={},
    xlabel={epochs},
    ylabel={$H(\Tilde{\alpha})$},
    xmin=-2, xmax=52,
    ymin=-0.1, ymax=3.1,
    tick align=outside,
    xtick={0, 10, 20, 30, 40, 50},
    ytick={0, 1, 2, 3},
    ymajorgrids=true,
    xmajorgrids=true,
    grid style=dotted,
    width=4.5cm,
    height=4.5cm,
    label style = {font=\scriptsize},
    legend style={nodes={scale=0.4, transform shape}},
    tick label style = {font=\tiny},
    every major tick/.append style={thick, major tick length=2pt, black},
    xtick pos=left,
    ytick pos=left,
    legend image post style={scale=0.5}
]
\addplot[
        color=darkgreen,
        mark=*,
        mark options={solid},
        mark size=1pt
        ]
        coordinates {
        (0,3) (10, 3) (20, 3) (30, 3) (40, 3) (50, 3)
        };
    \addplot[
        color=orange,
        mark=*,
        mark options={solid},
        mark size=1pt
        ]
        coordinates {
        (0,3) (5,2.85) (10, 2.8) (20, 2.77) (30, 2.65) (40, 2.5) (50, 2.52)
        };
    \addplot[
        color=red,
        mark=*,
        mark options={solid},
        mark size=1pt
        ]
        coordinates {
        (0,3) (5,2.15) (10, 2) (20, 1.8) (30, 1.9) (40, 1.8) (50, 1.7)
        };
    \addplot[
        color=blue,
        mark=*,
        mark options={solid},
        mark size=1pt
        ]
        coordinates {
        (0,3) (5,1.4) (10, 1.3) (20, 0.8) (30, 0.5) (40, 0.55) (50, 0.3)
        };
    \addplot[
        color=gray,
        style=dashed
        ]
        coordinates {
        (0,2.35) (50, 2.35)
        };

\end{axis}
\end{tikzpicture}
\end{subfigure}%

\caption{\scriptsize mIoU score on PascalVOC2012 \textit{val} set (left image) and $\Tilde{\alpha}$ entropy values (right image) for different values of the momentum rate parameter $\gamma$. $\gamma{=}0$ denotes no update of  $\Tilde{\alpha}_0$. For $\gamma$ reasonably small, we approach the entropy of ground-truth area distribution of categories (dashed line in the right image).}
\label{fig:miou_gamma}
\end{figure}

\noindent
{\bf Optimal Transport.} During early training phases, the MCE optimisation localises the relevant patches to each category. Though, initially, the actual area estimation is quite inaccurate. A low value of $\gamma$ ensures that inaccurate estimations are discarded while encouraging matching predictions to the initial uniform distribution. 
While training progress, $\gamma$ trades off between the current and past estimations. As a consequence of these observations, we set $\gamma$ to $0.02$ for all experiments unless explicitly stated.
As the network area predictions get more accurate, $\gamma$ becomes irrelevant. Indeed, as shown in Figure \ref{fig:alpha_plot}, the EMA update of $\Tilde{\alpha}$ stops after $\sim40$ epochs.
In Figure \ref{fig:miou_gamma} (left), we report the model's performance as mIoU$\%$ with varying values for $\gamma$. In practice, the slow EMA update improves $\approx 6 \%$ mIoU over the frozen uniform distribution case. For higher values of $\gamma$, the network predictions are vulnerable to its own bias favouring the most frequent of categories. Figure \ref{fig:miou_gamma} (right) shows that large $\gamma$s drive the network to a degenerate solution, exemplified by the entropy dropping to zero.
In \Cref{tab:abl}, we show that removing the OT step and relying only on global-local correspondence leads to a sharp drop in mIoU accuracy. 

\noindent
{\bf Match-loss.} We evaluate the importance of trading predictions and transportation plans between the network branches via a cross-entropy loss. In order to do this, we remove the connection between the two branches provided by Equation \ref{eq:loss} and minimise the cross-entropy between the predictions of a branch and its transportation plan. 
In \Cref{tab:abl}, we can see a substantial decline of the mIoU accuracy of $>10\%$, which we ascribe to the drop of the \textit{local-global} correspondence in the loss function.

\noindent
{\bf ViT backbone.} In \Cref{tab:abl} a comparison between different ViT architectures \cite{dosovitskiy2021image} performance is presented. While the accuracy of the $base$ and $large$ model are comparable, there is a   drop ($\approx 5-7\%$) for the $small$ version of ViT.
We deduce that the quality of the extracted features depends weakly on the model size.

\begin{table}[!htb]
\begin{minipage}{0.48\columnwidth}
    \centering
   \resizebox{0.8\columnwidth}{!}      
     {%
       \begin{tabular}{|c | c | c | c | }
         \multicolumn{4}{c}{}\\
         \hline 
         \rowcolor[gray]{.85}
         Backbone &  OT & Match &  Results    \\ 
         \hline
         \hline
         ViT-S/16 & $\times$ & \checkmark  & 35.12 \\
         & \checkmark &  $\times$ & 61.42 \\
         & \checkmark & \checkmark & 69.05 \\
         ViT-B/8 & \checkmark &  \checkmark & 73.12 \\
         ViT-B/16 & \checkmark &  \checkmark & \textbf{75.56} \\
         ViT-L/16 & \checkmark &  \checkmark & 75.24 \\
         \hline
     \end{tabular}}
    \caption{\scriptsize mIoU(\%) score for different backbone configurations on PascalVOC2012 train set.}
    \label{tab:abl}
\end{minipage}\hfill
\begin{minipage}{0.48\columnwidth}
    \centering
      \resizebox{0.96\columnwidth}{!}
     {%
       \begin{tabular}{|c | c | c | c | }
         \multicolumn{4}{c}{}\\
         \hline 
         \rowcolor[gray]{.85}
         Supervision & Method &  PascalVOC2012 &  COCO2014    \\ 
         \hline
         \hline
         WSSS & \textit{val} &  75.0  & 46.0 \\
         & \textit{test} & 75.7 & \\
         \hline
         USS & \textit{val} &  43.6  & 19.55 \\
         \hline
     \end{tabular}
     }
     \caption{\scriptsize mIoU($\%$) score for different supervision strategies on PascalVOC2012 and COCO2014.}
     \label{tab:results}
\end{minipage}
\end{table}
\noindent
{\bf Exploring Different Levels of Supervision.}
As shown in \Cref{tab:results} there is a drop of $\sim$30 points between weakly supervised and unsupervised mIoU. In this section, we examine how the availability of ground-truth labels influences these results. 
We randomly replace a fraction $\beta{\in}[0,1]$ of the ground-truth labels for the generated pseudo-labels, using Hungarian matching \cite{kuhn1955hungarian} to map one to the other. 
In \Cref{fig:true_pseudo}, we recognize an initial sharp drop in accuracy with $\beta {\in} [0, 0.2]$.
The drop indicates that the amount of the pseudo-labels strongly influences the model accuracy in this regime; with $\beta {\in} [0.2, 1.0]$, there is a smoother accuracy drop, indicating that in this interval, there might be some other component influencing this behaviour.
\begin{figure}[h!]
\centering
    \begin{minipage}{0.48\columnwidth}
    \centering
    \begin{tikzpicture}
    \begin{axis}[
    title style={},
    xlabel={$\beta$ (pseudo-labels mixing)},
    ylabel={score (\%)},
    xmin=-0.05, xmax=1.05,
    ymin=37, ymax=103,
    tick align=outside,
    xtick={0, .2, .4, .6, .8, 1},
    ytick={40, 60, 80, 100},
    legend pos=north east,
    ymajorgrids=true,
    xmajorgrids=true,
    grid style=dotted,
    width=4.5cm,
    height=4.5cm,
    label style = {font=\scriptsize},
    legend style={nodes={scale=0.4, transform shape}},
    tick label style = {font=\tiny},
    every major tick/.append style={thick, major tick length=2pt, black},
    xtick pos=left,
    ytick pos=left,
    legend image post style={scale=0.5},
]
\addplot[
    color=blue,
    mark=*,
    mark options={solid},
    mark size=1pt
    ]
    coordinates {
    (0.0,72) (0.1, 66) (0.2, 59) (0.6, 47) (1.0, 43)
    };
    \addlegendentry{mIoU}
\addplot[
    color=orange,
    mark=*,
    mark options={solid},
    mark size=1pt
    ]
    coordinates {
    (0.0,100) (0.1, 95) (0.2, 82) (0.6, 67) (1.0, 51)
    };
    \addlegendentry{F1 micro}
\addplot[
    color=darkgreen,
    mark=*,
    mark options={solid},
    mark size=1pt
    ]
    coordinates {
    (0.0,100) (0.1, 94) (0.2, 80) (0.6, 62) (1.0, 53)
    };
    \addlegendentry{F1 macro}
\end{axis}
\end{tikzpicture}
\caption{\scriptsize mIoU score on PascalVOC2012 \textit{val} set with respect to $\beta$-mixed class-labels quality, expressed as F1 score with respect to ground truth labels. The task is weakly supervised for $\beta{=}0$, unsupervised for $\beta{=}1$.}
\label{fig:true_pseudo}
\end{minipage}
\hfill
  \begin{minipage}{0.48\columnwidth}
\centering
    \begin{tikzpicture}
    \begin{axis}[
    title style={},
    xlabel={epochs},
    ylabel={distance ({\tiny${\times}10^{{-}1}$})},
    xmin=-2, xmax=52,
    ymin=-0.01, ymax=0.31,
    tick align=outside,
    xtick={0, 10, 20, 30, 40, 50},
    ytick={0, 0.1, 0.2, 0.3},
    legend pos=north east,
    ymajorgrids=true,
    xmajorgrids=true,
    grid style=dotted,
    width=4.5cm,
    height=4.5cm,
    label style = {font=\scriptsize},
    legend style={nodes={scale=0.4, transform shape}},
    tick label style = {font=\tiny},
    every major tick/.append style={thick, major tick length=2pt, black},
    xtick pos=left,
    ytick pos=left,
    legend image post style={scale=0.5},
]
\addplot[
    color=blue,
    mark=*,
    mark options={solid},
    mark size=1pt
    ]
    coordinates {
    (0,0.127) (10,0.06) (20, 0.041) (30, 0.03) (40, 0.025) (50, 0.023)
    };
    \addlegendentry{$D_{JS}(\Tilde{\boldsymbol\alpha}_{m-1}{\parallel} \Tilde{\boldsymbol\alpha}_m)$}
\addplot[
    color=red,
    mark=*,
    mark options={solid},
    mark size=1pt
    ]
    coordinates {
    (0,0.29) (10, 0.13) (20, 0.083) (30, 0.067) (40, 0.060) (50, 0.057)
    };
    \addlegendentry{$D_{JS}(\boldsymbol\alpha^*{\parallel} \Tilde{\boldsymbol\alpha}_{m})$}
\end{axis}
\end{tikzpicture}
\caption{\scriptsize Evolution of the EMA update for $\gamma{=}0.02$. The blue line shows the distance of $\Tilde{\boldsymbol\alpha}_{m}$  form $\Tilde{\boldsymbol\alpha}_{m{-}1}$ at epoch $m$. The red line shows $\Tilde{\boldsymbol\alpha}_{m}$ convergence to ground-truth normalized area distribution of categories $\boldsymbol\alpha^\star$.  }
\label{fig:alpha_plot}
  \end{minipage}
\end{figure}

\begin{figure*}[bt!]
\centering
 \includegraphics[width=0.99\linewidth]{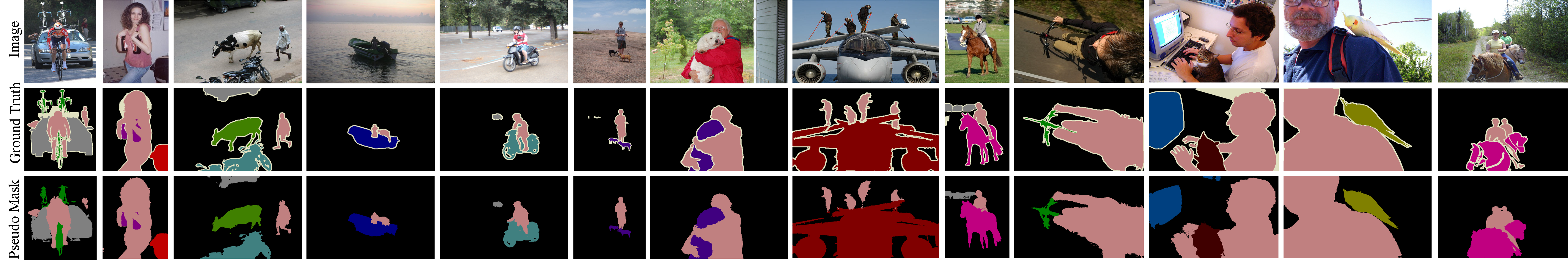}
\caption{Qualitative results for Weakly-Supervised Semantic Segmentation  on  PascalVOC2012: here the mask are predicted with the supervision of the given image-class labels.}
\label{fig:qualit_wsss}
\end{figure*}%

\begin{figure*}[bt!]
\centering
 \includegraphics[width=0.99\linewidth]{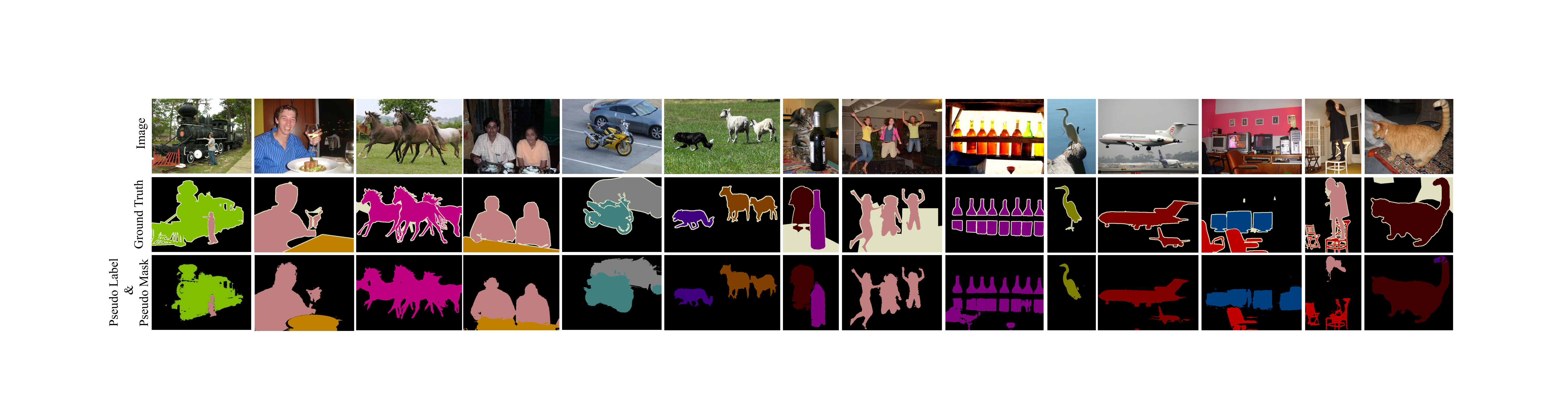}
\caption{Qualitative results on  PascalVOC2012 for Unsupervised Semantic Segmentation: here both the image-class labels and the masks are predicted with no supervision. }
\label{fig:qualit_coco}
\end{figure*}%

\begin{figure*}[bt!]
\centering
 \includegraphics[width=0.99\linewidth]{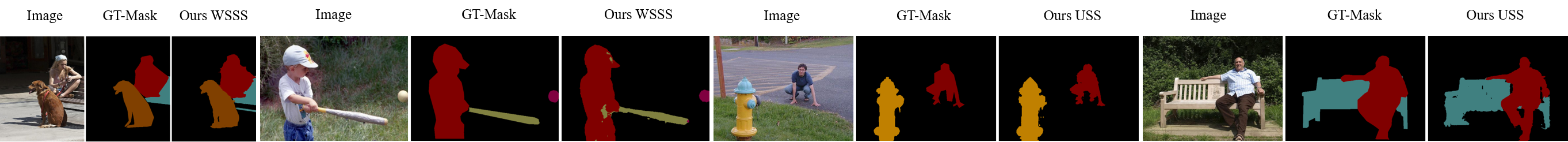}
\caption{Qualitative results on  MS-COCO2014 for weakly and unsupervised segmentation.}
\label{fig:qualit_unsup}
\end{figure*}%
\noindent
\subsection{Qualitative results}
We show our qualitative results in Figures \ref{fig:qualit_wsss} to \ref{fig:qualit_unsup} for the weakly supervised and unsupervised tasks. Note that the white regions in the ground-truth masks are no-class. Our PC2M-generated pseudo-masks exhibit close-to-ground-truth mask shapes and contours. 
In Appendix \ref{sec:Unsup_method} more qualitative results on both WSSS and USS on PascalVOC2012 and on MS-COCO2014 datasets can be found, together with qualitative comparisons to other approaches.

\typeout{---------------Comparisons --------------------}
\subsection{Comparison to the State of the Art} 
\noindent
{\bf Weakly supervised semantic segmentation}
We compare our method with the most recent approaches in the literature (2021-2022) on PascalVOC2012\cite{everingham2010pascal}, and MS-COCO2014 \cite{lin2014microsoft} since mIoU scores steadily grow over time. We have used the same nomenclature of ViT-PCM \cite{RossettiECCV-2022} distinguishing between raw Baseline Pseudo-Masks (BPM) and those refined with CRF \cite{krahenbuhl2011efficient}. See columns 3-4 of \Cref{tab:wss1}, where the contribution of CRF is apparent. On validation and test, DeepLab \cite{Chen2018DeepLabSI} is used to verify the predicted pseudo-masks, often resorting to a different backbone than the one used for predicting the pseudo-masks. Note that, as indicated with the ${\star}$ in \Cref{tab:wss1,tab:wss2,tab:wss3,tab:ussco,tab:usspv}, we do not train any additional segmentation network to improve generalization, such as DeepLab, nor we use additional forms of supervision, such as saliency maps. 
Several authors \cite{araslanov2020single,ru2022learning,RossettiECCV-2022} have observed that there is a striking difference between end-to-end networks and multi-stage networks.
We can appreciate the disparity by looking at the accuracy jump between \Cref{tab:wss1} and \Cref{tab:wss2} for some models and the number of parameters used, for example, with Resnet38. 

Just a few approaches have been tested on MS-COCO2014, due to the larger number of classes and resources required. 
Good results are attained by AFA \cite{ru2022learning}, MCTformer\cite{xu2022multi} (which has an end-to-end version), and ViT-PCM \cite{RossettiECCV-2022}, all based on ViT \cite{dosovitskiy2021image}.  We also outperform the state-of-the-art on MS-COCO2014.

  \begin{table}[h]
     \centering
     \resizebox{0.98\columnwidth}{!}
     {%
       \begin{tabular}{|c | c | c | c | } 
        \multicolumn{4}{c}{}\\
        \hline
        \rowcolor[gray]{.85}
         Method &  Backbone & BPM             & BPM{+}CRF \\ 
         \hline\hline
             AdvCAM\cite{lee2021anti}\tiny{\textsc{CVPR'21} } & ResNet50 & 55.60 &  62.10 \\ 
             CPN\cite{zhang2021complementary}\tiny{\textsc{ICCV'21} } & ResNet38 &  57.43 & - \\
             CSE\cite{kweon2021unlocking}\tiny{\textsc{ICCV'21} } & ResNet38 &56.0 & 62.8\\
             EDAM\cite{wu2021embedded}\tiny{\textsc{CVPR'21} } &  ResNet101 &  52.83 &  58.18 \\
             AMR\cite{qin2022activation}
            \tiny{\textsc{AAAI'22} } &  Resnet50 &  56.8 & 69.7 \\  
             MCTformer\cite{xu2022multi}\tiny{\textsc{CVPR'22} } &  DeiT-S &  61.70 &  - \\
             PPC\cite{du2022weakly}\tiny{\textsc{CVPR'22} } & Resnet38 &  61.50 &  64.00\\
             CLIMS\cite{xie2022clims}\tiny{\textsc{CVPR'22} } & Resnet50 &  56.60 &  -\\
             SIPE\cite{chen2022self}\tiny{\textsc{CVPR'22} } & Resnet50 &  58.60 & 64.70\\            
             AdvCAM+W-OoD\cite{lee2022weakly}\tiny{\textsc{CVPR'22} } & Resnet50&  59.10 & 65.50\\
            AFA\cite{ru2022learning}\tiny{\textsc{CVPR'22} } & MiT-B1 &  63.80 & 66.00\\
             ViT-PCM\cite{RossettiECCV-2022}\tiny{\textsc{ECCV'22}} & ViT-B/16 &  67.71 &  71.4 \\
             {\bf Ours}$^\star$ & ViT-B/16  &  \textbf{75.56} &  {\bf 76.55} \\
             \hline
    \end{tabular}}
    \caption{WSSS methods mIoU(\%) on PascalVOC2012 \textit{train} set. The $\star$ indicates we do not train a second segmentation network.}
    \label{tab:wss1}
    \end{table}
   
 \begin{table}[h]
   \centering
         \resizebox{0.99\columnwidth}{!}
         {%
         \begin{tabular}{|c | c | c | c | }
          \multicolumn{4}{c}{}\\
              \hline
              \rowcolor[gray]{.85}
              Method &  Backbone &      Val & Test \\ 
             \hline\hline
             AdvCAM\cite{lee2021anti}\tiny{\textsc{CVPR'21} } & ResNet50 & 68.1 &  68.0 \\ 
             CPN\cite{zhang2021complementary}\tiny{\textsc{ICCV'21} } & ResNet38 &  67.8 & 68.5 \\
             CSE\cite{kweon2021unlocking}\tiny{\textsc{ICCV'21} } & ResNet38 &68.4 & 68.2\\
             EDAM\cite{wu2021embedded}\tiny{\textsc{CVPR'21} } &  ResNet101 &  52.83 &  58.18 \\
             AMR\cite{qin2022activation}
            \tiny{\textsc{AAAI'22} } &  Resnet101 &  68.8 & 69.1 \\  
             \cite{wang2022looking}\tiny{\textsc{CVPR'22} } &  Resnet101 &  66.2 & 66.9 \\
             MCTformer\cite{xu2022multi}\tiny{\textsc{CVPR'22} } &  Resnet38 &  71.9 & 71.6 \\
             PPC\cite{du2022weakly}\tiny{\textsc{CVPR'22} } & Resnet38 &  72.60 &  73.60\\
             CLIMS\cite{xie2022clims}\tiny{\textsc{CVPR'22} } & Resnet50 &  70.4 &  70.0\\
             SIPE\cite{chen2022self}\tiny{\textsc{CVPR'22} } & Resnet101 &  68.8 & 69.7\\
             AdvCAM+W-OoD\cite{lee2022weakly}\tiny{\textsc{CVPR'22} } & Resnet38 &  70.7 & 70.1\\
             MCIS\cite{sun2020mining}\tiny{\textsc{ECCV'20} }& ResNet101 &66.2 & 66.9\\
            AFA\cite{ru2022learning}\tiny{\textsc{CVPR'22} } & MiT-B1 &  66.0 & 66.3\\
             ViT-PCM\cite{RossettiECCV-2022}\tiny{\textsc{ECCV'22}}& ResNet 101 &  70.3 &   70.9 \\
             {\bf Ours}$^{\star}$ & ViT-B/16 & {\bf 75.0}    &  {\bf 75.7}  \\
         \hline  
          \end{tabular}
        }
    \caption{mIoU(\%) of WSSS methods on PascalVOC2012 \textit{val} and \textit{test} set. The $\star$ indicates we do not train a second segmentation network.}
    \label{tab:wss2}
  \end{table}
  \begin{table}[h]
     \centering
       \resizebox{0.71\columnwidth}{!}
     {%
       \begin{tabular}{|c | c | c |  }
         \multicolumn{3}{c}{}\\
         \hline 
         \rowcolor[gray]{.85}
         Method &  Backbone &  {Val}    \\ 
         \hline
         \hline
         RIB\cite{lee2021reducing}\tiny{\textsc{Neurips'2021} }& R101 & 43.8\\ 
         MCTformer\cite{xu2022multi}\tiny{\textsc{CVPR'22} } &  Resnet38 &  {42.0} \\
          SIPE\cite{chen2022self}\tiny{\textsc{CVPR'22} } & Resnet38 &  {43.6} \\
          AFA\cite{ru2022learning}\tiny{\textsc{CVPR'22} } & MiT-B1 &  38.9\\
          ViT-PCM\cite{RossettiECCV-2022}\tiny{\textsc{ECCV'22} }  & ViT-B/16 &  45.0\\
          {\bf Ours}$^\star$ & ViT-B/16 & {\bf 46.0}\\
          \hline

     \end{tabular}}
    \caption{mIoU(\%) of WSSS methods on MS-COCO2014 \textit{val} set. The $\star$ indicates we do not train a second segmentation network.}
    \label{tab:wss3}
\end{table}

  \begin{table}[h]
     \centering
       \resizebox{0.99\columnwidth}{!}
    {%
       \begin{tabular}{|c | c | c | c| c| }
         \multicolumn{3}{c}{}\\
         \hline 
         \rowcolor[gray]{.85}
         Method & Pretrain Dataset &Backbone & Initial Weights &  {Val}    \\ 
         \hline
         \hline
         MaskContrast\cite{van2021unsupervised}\tiny{\textsc{CVPR'2021} }& COCO & - & MoCo-v2\cite{chen2020improved} & 35.0\\  
         Leopart\cite{ziegler2022self}\tiny{\textsc{CVPR'2022} }& COCO& ViT-S/16& DINO\cite{caron2021emerging} & 41.7\\  
         SemSpectral\cite{melas2022deep}\tiny{\textsc{CVPR'2022} }& ImageNet & - & DINO\cite{caron2021emerging} & 30.8
         \\
         HSG\cite{ke2022unsupervised}\tiny{\textsc{CVPR'2022} }& COCO & ResNet 50 & - & 41.9\\
         TransFGU\cite{yin2022transfgu}\tiny{\textsc{ECCV'2022} }& ImageNet & ViT-S/8 & DINO\cite{caron2021emerging} & 37.15\\
          {\bf Ours}$^\star$ & ImageNet & ViT-B/16 & DINO\cite{caron2021emerging} & {\bf 43.6}\\
          \hline

     \end{tabular}}
     \caption{mIoU(\%) on Unsupervised PascalVOC2012 \textit{val}  set. The $\star$ indicates we do not train a second segmentation network.}
     \label{tab:usspv}
\end{table}
\begin{table}[h]
     \centering
       \resizebox{0.99\columnwidth}{!}
    {%
       \begin{tabular}{|c | c | c | c| c| }
         \multicolumn{3}{c}{}\\
         \hline 
         \rowcolor[gray]{.85}
         Method & Pretrain Dataset &Backbone & Initial Weights &  {Val}    \\ 
         \hline
         \hline
         MaskContrast\cite{van2021unsupervised}\tiny{\textsc{CVPR'2021} }& COCO & - & MoCo-v2\cite{chen2020improved} & 3.73\\  
         TransFGU\cite{yin2022transfgu}\tiny{\textsc{ECCV'2022} }& ImageNet & ViT-S/8 & DINO\cite{caron2021emerging} & 12.69\\
          {\bf Ours}$^\star$ & ImageNet & ViT-B/16 & DINO\cite{caron2021emerging} & {\bf 19.55}\\ 
          \hline

     \end{tabular}}
     \caption{mIoU(\%) of Unsupervised Semantic Segmentation on MS-COCO2014 \textit{val}  set. The result of MaskContrast has been reproduced  by\cite{yin2022transfgu}. The $\star$ indicates we do not train a second segmentation network.}
     \label{tab:ussco}
\end{table}

\noindent
{\bf Unsupervised semantic segmentation}. We compare our method with all available state-of-the-art results on PascalVOC2012 and MS-COCO2014; see \Cref{tab:usspv} and \Cref{tab:ussco}. Initial methods confronted with COCO-stuff \cite{caesar2018coco}  and subsets of COCO-stuff due to the difficulty of dealing with the background. The methodology in \cite{van2021unsupervised,melas2022deep} uses weights from self-supervised representation learning, trained on ImageNet and further applies the unsupervised clustering method. Finally, DeepLab \cite{Chen2018DeepLabSI} is trained with a backbone pre-trained with self-supervision, and the output is matched with ground-truth labels from the validation set via the Hungarian matching \cite{kuhn1955hungarian}. 
On the other hand,  HSG   \cite{ke2022unsupervised}, Leopart\cite{ziegler2022self} and TransFGU\cite{yin2022transfgu} evaluate directly on the predicted pseudo-mask. We do not use DeepLab either,  outperforming current state-of-the-art; by the way, DeepLab results can be found in the Appendix for completeness.

On MS-COCO2014, our only competitor is TransFGU \cite{yin2022transfgu}; the reported results for MaskContrast are from the TransFGU re-implementation with ViT backbone trained by DINO. We improve their methods by 6.86 points in mIoU. 

\typeout{----------------Qualitative Results --------------------}

\typeout{----------------Limitations --------------------}
\section{Limitations}\label{sec:limit}
Our PC2M models a categorical distribution over patches, so using it as a plug-and-play solution in CAM-based methods can be challenging. Indeed, a mapping of the CAM activations to the categorical distribution would be necessary.
While for WSSS, the improvement has touched 15 points in mIoU in the last two years, the gap of USS  with WSSS is $\sim$30 points in mIoU. Knowledge of contextual information is crucial to our top-down framework, and for this reason, it should be improved. Future work requires moving the pseudo labels generation in the loop with the mask prediction, especially caring about multi-labelling.

\typeout{----------------Conclusions --------------------}
\section{Conclusions}\label{sec:concl}
We presented the PC2M that proves excellent performance on semantic segmentation supervised by image-class labels only. We extended the method to unsupervised semantic segmentation by predicting pseudo-labels that can capture the multiplicity of objects in datasets such as PascalVOC2012 and MS-COCO2014. PC2M uses pseudo-labels to predict pseudo-masks. The gap between unsupervised semantic segmentation and the weakly supervised one remains relevant. Nevertheless, the results are promising from the perspective of conjugating the pseudo-labels prediction step with the pseudo-masks one. More details are in the Appendices. 

\newpage
\clearpage

\begin{appendices}

\typeout{-------------- Intro ---------------------}
\noindent
In these appendices, we make additional considerations on the PC2M model introduced in the main paper,  providing  further results in the following contents.
 
\typeout{-------------- Convergence ---------------------}

\section{Considerations on the discrete metric $\boldsymbol\alpha$ }\label{sec:dist_a}
\noindent
In the main paper, we considered the optimal transport problem with entropic regularization \cite{peyre2019computational}:
\begin{equation}\label{eq:OT_}
\displaystyle{
L^{\epsilon}_S({\boldsymbol \delta}, {\boldsymbol \alpha}) = \min_{Q\in U({\boldsymbol \delta}, {\boldsymbol \alpha})} tr(Q^{\top}S) -\epsilon H(Q) }
\end{equation}
\noindent
for both the branches of the network, with costs  $S^{\ell}$ and $S^g$. 
Note that the costs $S^{\ell} {=}{-}\log(P^{\ell})$ and $S^{g} {=}{-}\log(P^{g})$ are both convex and bound. In fact $0{<} P^{\ell},P^{g}{<} 1$.

\noindent 
Nevertheless, the proposed costs are not metrics, which means that, in our case, we cannot consider OT a metric. Our objective  is specifically on the transportation plans $Q^{\ell}$ and $Q^{g}$, which we treat as pseudo-labels in the loss ${\mathcal L}_{match}$  to constrain the  network predictions.

\noindent
We have specified the average area distribution $\tilde{\boldsymbol\alpha}$ as :
\begin{equation}\label{eq:alpha}
\begin{array}{l}
\tilde{\boldsymbol\alpha_0}^{\ell} {=}\tilde{\boldsymbol\alpha_0}^{g} {=}\boldsymbol{\nu}_D
\mbox{ and }\\
\Tilde{\alpha}_m^c {=}  (1{-}\gamma) \Tilde{\alpha}_{m{-}1}^c  {+} \gamma \frac{1}{T}\sum_{X{\in}D} q_{m-1}^c(X)
\end{array}
\end{equation}
with $q_m^c(X){=}\displaystyle{\frac{1}{K}\sum_{k=1}^{K} p(y_k{=}c| u_k,F,W^c)}$
 
\noindent
Thus we have two optimal transport problems:
\begin{equation}\label{eq:OT_2}
\displaystyle{
L^{\epsilon}_{S^{\ell}}({\boldsymbol \delta}, {\boldsymbol \alpha}^{\ell}) \mbox{ and }
L^{\epsilon}_{S^{g}}({\boldsymbol \delta}, {\boldsymbol \alpha}^{g})}  
\end{equation}
\noindent
We show that:\\
1) $\tilde{\boldsymbol\alpha}^{\ell} = \tilde{\boldsymbol\alpha}^g$.  

\noindent
2) $Q^{\ell} \neq Q^g$.

\noindent
3) If $\tilde{\alpha}^c_m = \tilde{\alpha}^c_{m-1}$ then
 \begin{equation}\label{eq:converg2_}
\Tilde{\alpha}_m^c = \Tilde{\alpha}_{m{-}1}^c = \frac{1}{T}\sum_{X{\in}D} q_{m-1}^c(X) = \frac{1}{T{\cdot} K}\sum_{j\in T\cdot K} P_j^c.
\end{equation}

\noindent
We first prove 1). 
\begin{proof}
We recall $\tilde{\boldsymbol\alpha}$ is computed off-line at each epoch on the dataset current predictions.
With a straightforward induction, since $\tilde{\boldsymbol\alpha}_0^{\ell} {=} \tilde{\boldsymbol\alpha}_0^g {=}\boldsymbol{\nu}_D$, and since $\gamma$ is fixed, we have that $\tilde{\boldsymbol\alpha}^{\ell}_{m{-}1} {=}\tilde{\boldsymbol\alpha}_{m{-}1}^g$.  
\noindent
By  definition,  $q_{m}^c(X)$  is the prediction map for $X$, namely, the probability for any patch of the input image $X$ to belong to class $c$, $c{\in} C$, obtained as  the output prediction of the network, via ${\mathcal L}_{PC2M}$, at the current epoch $m$. And here, we are considering all images in the dataset $D$. It follows that $q_{m}^c(X)$ is equal for both  branches, for all $c\in C$. Therefore $\tilde{\boldsymbol\alpha}_{m}^{\ell} {=} \tilde{\boldsymbol\alpha}_{m}^g$ for all $m$. 
\end{proof}

\noindent
Next, we prove 2). 
\begin{proof}
Suppose $Q^{\ell} = Q^{g}$ and they both are transport plans satisfying Eq.~\ref{eq:OT_}. Then, according to Eq.~\ref{eq:OT_}, it must be that 
$(Q^{\ell})^{\top}S^{\ell} {=} (Q^g)^{\top}S^g$, and therefore that $S^{\ell} {=} S^g$.
Now  $S^{\ell} {=} {-}\log(P^{\ell})$ and $S^{g} {=} {-}\log (P^{g})$, with these costs fixed, and $P^{\ell}$ and $P^g$ are the predictions from the two branches of the network, under the hypothesis that $P^{\ell}$ and $P^g$ are different, due to different augmentations in the two branches. It follows that $S^{\ell} {\neq} S^g$, hence the $Q^{\ell}$ and $Q^g$ minimizing Eq. \ref{eq:OT_} must be such that $Q^{\ell} {\neq} Q^{g}$.
\end{proof}

\noindent
Here we prove 3). 
\begin{proof}
Suppose $\tilde{\alpha}^c_m {=}\tilde{\alpha}^c_{m-1}$,  by simple manipulations of Eq. \ref{eq:alpha}, it follows that $\tilde{\alpha}^c_m {=} \frac{1}{T}\sum_{X{\in}D} q_{m-1}^c(X)$. Here $c$ is any $c{\in} C$.

\noindent
By definition of optimal transport and by 1) above, this implies that $\tilde{\boldsymbol\alpha}_m$ is the marginal of $Q^{\ell}$ and also the marginal of $Q^{g}$. 
On the other hand,  $P_j^c$ is the estimated probability of the patches at the network output via ${\mathcal L}_{PC2M}$, therefore
$\Tilde{\alpha}_{m}^c {=} \frac{1}{T}\sum_{X{\in}D}q_{m-1}^c(X) {=} \frac{1}{T{\cdot} K}\sum_{j{\in} T{\cdot} K} P_j^{c}$.

\noindent
Since $\frac{1}{T{\cdot} K}\sum_{j\in T\cdot K} P_j^{c}$ is the network output, by the fact that the marginal of $P$ is equal to the  marginal of both $Q^{\ell}$ and $Q^g$ it follows that at the minimum of the relative entropy\footnote{Here we used the definition as given in \cite{Cover2006}.},  also the marginals of $P^{\ell}$ and $P^{g}$ are both $\tilde{\boldsymbol\alpha}_m$.  
As shown in the main paper, since the cost $S^{\ell}{=} {-}\log\left(P^{\ell}\right)^{1{/}\epsilon} $ and the cost $S^{g}{=} {-}\log\left(P^{g}\right)^{1{/}\epsilon}$, the Sinkhorn algorithm stops updating. The 
update, for $t{\in}\{g,l\}$, is $\mbox{diag}({\bf u}) (P^{t})^{1/\epsilon}\mbox{diag}({\bf v}) ({\mathds 1}_{|C|}){=} {\boldsymbol \delta}$ and $\mbox{diag}({\bf v})[(P^{t})^{1/\epsilon}]^{\top}\mbox{diag}({\bf u})({\mathds 1}_{TK}) {=} {\boldsymbol \tilde{\alpha}_m}$ which is already satisfied  for the unknown variables $({\bf u}, {\bf v}){=} ({\mathds 1}_{TK},{\mathds 1}_{|C|})$, and  the hypothesis that $\epsilon {\approx}1$ towards the end of training. \end{proof}

\subsection{Empirical convergence of {\larger $\alpha$} to ground truth average areas}\label{sub-sec:empirical}
Given the above statements, it is natural to ask whether the normalised area $\tilde{\boldsymbol\alpha}$ converges to the ground truth normalised area
 $\boldsymbol{\alpha}^{*} {=} (\frac{1}{T} \sum_{j\in T} A_j^{c},\ldots,\frac{1}{T} \sum_{j\in T} A_j^{|C|})^{\top}$, where $A^c$ is the ground truth area of category $c$. We 
 show experimentally that $\|\tilde{\boldsymbol\alpha}{-}\boldsymbol{\alpha}^{*}\|{<}\lambda$.

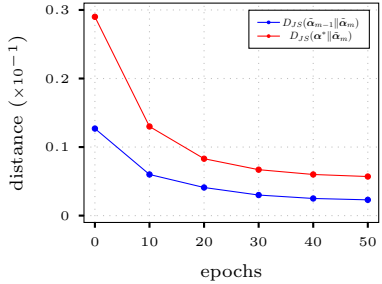
\begin{figure}[t]
    \centering
    \begin{tikzpicture}
    \begin{axis}[
    title style={},
    xlabel={epochs},
    ylabel={distance ({\tiny${\times}10^{{-}1}$})},
    xmin=-2, xmax=52,
    ymin=-0.01, ymax=0.31,
    tick align=outside,
    xtick={0, 10, 20, 30, 40, 50},
    ytick={0, 0.1, 0.2, 0.3},
    legend pos=north east,
    ymajorgrids=true,
    xmajorgrids=true,
    grid style=dotted,
    width=5.5cm,
    height=4.5cm,
    label style = {font=\scriptsize},
    legend style={nodes={scale=0.4, transform shape}},
    tick label style = {font=\tiny},
    every major tick/.append style={thick, major tick length=2pt, black},
    xtick pos=left,
    ytick pos=left,
    legend image post style={scale=0.5},
]
\addplot[
    color=blue,
    mark=*,
    mark options={solid},
    mark size=1pt
    ]
    coordinates {
    (0,0.127) (10,0.06) (20, 0.041) (30, 0.03) (40, 0.025) (50, 0.023)
    };
    \addlegendentry{$D_{JS}(\Tilde{\boldsymbol\alpha}_{m-1}{\parallel} \Tilde{\boldsymbol\alpha}_m)$}
\addplot[
    color=red,
    mark=*,
    mark options={solid},
    mark size=1pt
    ]
    coordinates {
    (0,0.29) (10, 0.13) (20, 0.083) (30, 0.067) (40, 0.060) (50, 0.057)
    };
    \addlegendentry{$D_{JS}(\boldsymbol\alpha^*{\parallel} \Tilde{\boldsymbol\alpha}_{m})$}
\end{axis}
\end{tikzpicture}
\caption{\scriptsize Evolution of the EMA update for $\gamma{=}0.02$. The blue line shows the distance of $\Tilde{\boldsymbol\alpha}_{m}$  form $\Tilde{\boldsymbol\alpha}_{m{-}1}$ at epoch $m$. The red line shows $\Tilde{\boldsymbol\alpha}_{m}$ convergence to ground-truth normalized area distribution of categories $\boldsymbol\alpha^\star$.  }
\label{fig:alpha_plot_}
\end{figure}

\noindent
Figure \ref{fig:alpha_plot_} shows the evolution during training of the estimated average area distribution of categories for the PascalVOC2012\cite{everingham2010pascal} \textit{trainaug} dataset \cite{hariharan2011semantic}, using $\gamma=0.02$ for EMA update (see Eq.~{\color{red} 7}  in the main paper). Namely, we use Jensen–Shannon divergence ($D_{JS}$) to assess the similarity between distributions. The red line in Figure \ref{fig:alpha_plot_} shows the divergence at epoch $m$ between $\tilde{\boldsymbol\alpha}_{m}$ and the ground truth area distribution of categories $\boldsymbol{\alpha}^*$. 
This show, experimentally, that $\|\tilde{\boldsymbol\alpha} {-}\boldsymbol{\alpha}^{*}\|{<}\lambda$, for $\lambda {<}0.05$. The blue line in Figure \ref{fig:alpha_plot_} shows the convergence of the slow-moving distribution $\Tilde{\boldsymbol\alpha}$ to a fixed target; that is, we compute the divergence at epoch $m$ between $\Tilde{\boldsymbol\alpha}_m$ and $\Tilde{\boldsymbol\alpha}_{m-1}$ to assess the magnitude of change of the distribution during the epochs. A similar graph can be shown between $\Tilde{\boldsymbol\alpha}_m$ and  $\frac{1}{T}\sum_{X{\in}D} q_{m-1}^c(X)$. 

\begin{figure*}[h]
    \centering
    \includegraphics[width=0.9\textwidth]{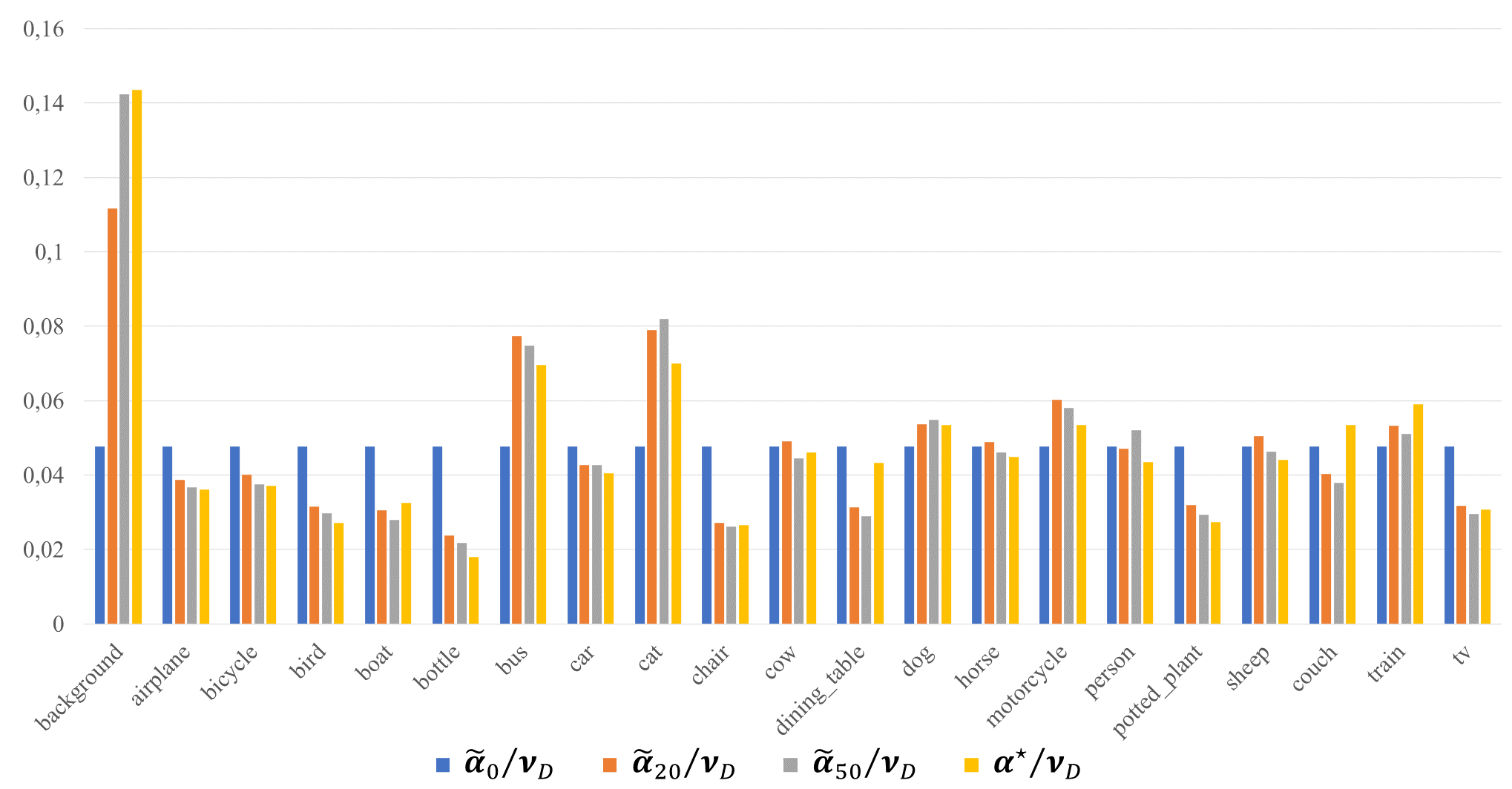}
    \caption{Temporal evolution of estimated unitary area distribution $\Tilde{\boldsymbol\alpha}_m/{\boldsymbol\nu_D}$ at epochs $m=\{0,20,50\}$ on PascalVOC2012 \textit{trainaug} set. $\Tilde{\boldsymbol\alpha}$ is the categories' area distribution, $\boldsymbol\nu_D$ is the relative class frequency, and $\boldsymbol\alpha^\star$ is the ground truth categories' area distribution in the dataset.}
    \label{fig:alpha_dist}
\end{figure*}

\noindent
In particular, as stated in Eq.~{\color{red} 5} in the main paper, $\boldsymbol\alpha$, which is used in Algorithm \ref{problem:algo}, is defined to be $\frac{\boldsymbol{\nu}_B}{Z\boldsymbol{\nu}_D}\tilde{\boldsymbol\alpha}$, with  $\boldsymbol{\nu}_B$ the relative frequency of categories in the batch, here equal to 1 to obtain the unitary area. Noticing that $\boldsymbol{\nu}_D$ is a constant, it follows that the above results also hold for $Z\boldsymbol\alpha$.
Therefore, in \Cref{fig:alpha_dist} we show that the above-assessed similarity is consistently extended to $\|\alpha^c{-}\alpha^{c,*}\|{<}\lambda$ up to a multiplicative factor proportional to  $Z$, for all $c\in C$.

\typeout{-------------- Alpha ---------------------}
\section{Algorithmic Details}\label{sec:alpha}
\noindent
We present our pseudocode in Algorithm \ref{problem:algo}. As stated in the main paper, we solve an optimal transport problem with regularised entropy penalty term to hasten convergence, which can be solved using a simple, alternate minimisation scheme. Since the objective is an $\epsilon$-strongly convex function, problem in Eq. \ref{eq:OT} has a unique optimal solution \cite{peyre2019computational}. To do so we instantiate the Sinkhorn-Knopp algorithm following considerations in \cite{peyre2019computational} and implementation in \cite{cuturi2013sinkhorn}. The number of iterations controls the trade-off between the optimality of the transport plan and computational cost. We empirically obtain the best compromise in terms of speed and performance, limiting the kernel balancing to 3 updates. A key insight is that, as $\epsilon{\rightarrow} {+}\infty$, the optimal coupling becomes less and less sparse, and the solution leads to the tensor product ${\boldsymbol \delta}\otimes{\boldsymbol \alpha}$. In contrast, as $\epsilon{\rightarrow} 0$, the unique solution converges to the maximum entropy solution of the standard OT problem, and the convergence of Sinkhorn’s algorithm deteriorates \cite{peyre2019computational}. We empirically found  that using $\epsilon{=}0.1$ works the best in our experiments when using three iterations only. Moreover, we increase the value of $\epsilon$ with an exponential growth of $0.05$ rate for 40 epochs until reaching the final value of $\epsilon{=}0.9$.
\begin{problem}\label{problem:algo}
\begin{figure}[h]
\centering
\includegraphics[width=0.99\linewidth]{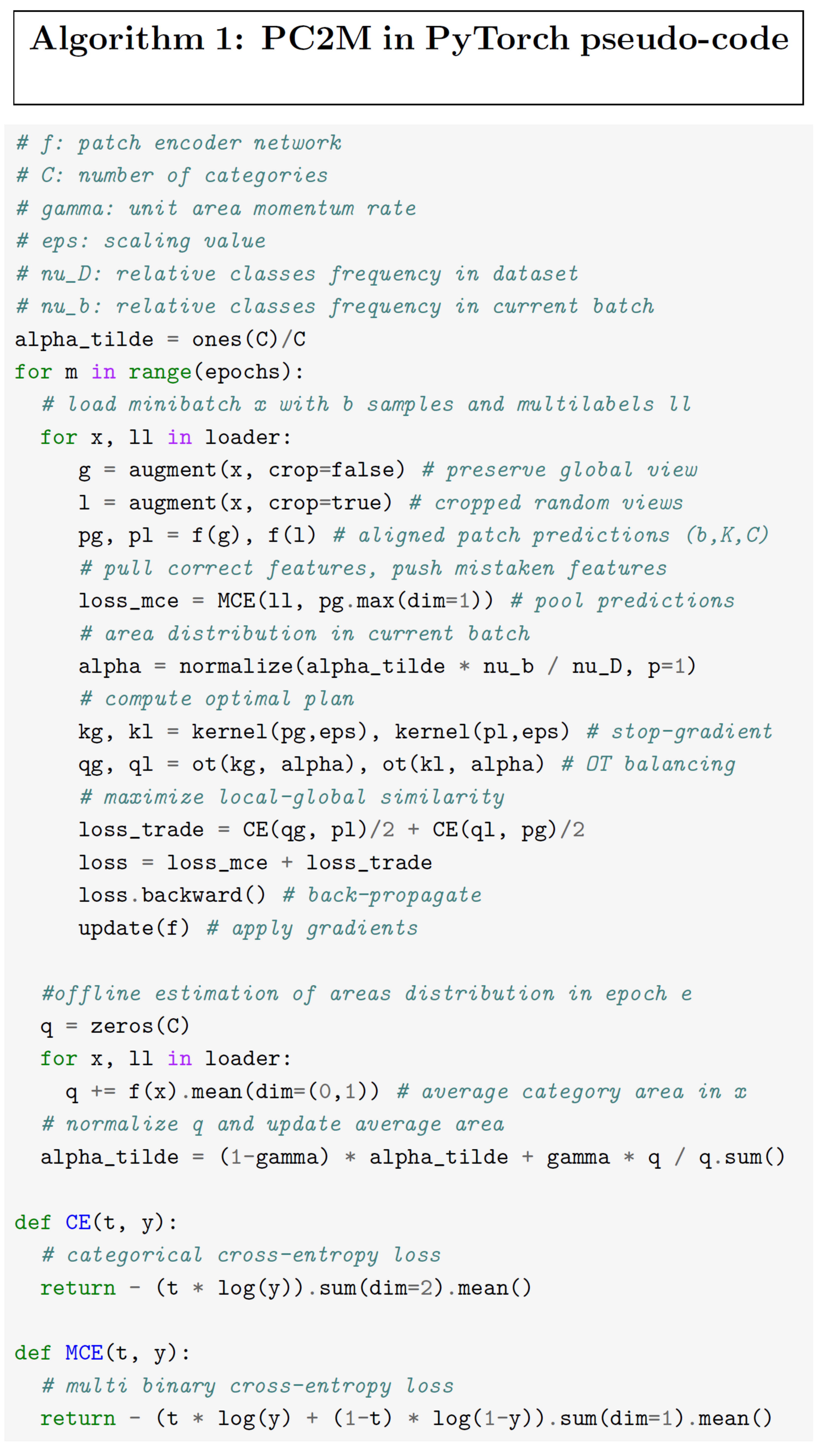}
\end{figure}
\end{problem}

\label{sec:other}

\typeout{-------------- Computational Cost --------------}
\section{Computational Cost}\label{sec:cost}
In this section, we analyse our framework's computational cost in terms of space and time complexity.
As highlighted in \cite{RossettiECCV-2022} ViT-based architectures have a memory footprint comparable with other weakly supervised segmentation models, as evidenced by the number of parameters of the model, which reaches 89.4 M. Regarding the running time of the model, we refer to Section 4.2 in the main paper. In \Cref{tab:par_size}, we show that the memory footprint of our unsupervised method is on par with other self-supervised frameworks. 
\begin{table}[h]
  \scriptsize
     \centering
       \resizebox{\columnwidth}{!}
     {%
       \begin{tabular}{|c | c | c | c|}
         \multicolumn{3}{c}{}\\
         \hline 
         \rowcolor[gray]{.95}
         Model & Architectures & \# Params ($\times 10^6$)  & mIoU\%  \\ 
         \hline
         \hline
         \textbf{Ours}  & ViT-B/16 +  & &\\
        & ViT-PCM \cite{RossettiECCV-2022} (ViT-B/16) &  175 & 43.6 \\
        &&&\\
         SemSpectral \cite{melas2022deep} & ViT-B/8 & &\\
        & DeepLab v3 (RN50) \cite{Chen2018DeepLabSI} (RN50) &  128 & 30.8 \\
        &&&\\
         MaskContrast \cite{gansbeke2021}& DeepLab v3 (DRN50) +  &  &\\
          & BASNet \cite{Qin_2019_CVPR}+&&\\ & DeepUSPS \cite{NEURIPS2019_54229abf} (DRN105) & 182 & 35.0\\
          &&&\\
         TransFGU \cite{yin2022transfgu} &  2*ViT-S/8 &&\\
         &+ 2*Segmenter \cite{strudel2021segmenter} (ViT-S/8) & 88 & 37.15 \\
         &&&\\
          Leopart \cite{ziegler2022self} & 2*ViT-B/8 & 172 & 41.7 \\
         \hline
     \end{tabular}}
     \caption{Backbone and number of parameters for the  state-of-the-art unsupervised segmentation models}
     \label{tab:par_size}
\end{table}

Consider the optimisation problem in \eqref{eq:OT} with $Q{\in}\mathbb{R}_+^{m{\times}n}$ and let $\eta{=}\max(m, n)$. For a large enough batch size, $\eta$ corresponds to the number of input patches.
Finding solutions to the OT problem, computed via the Sinkhorn-Knopp (SK) algorithm, at each training iteration adds some computational cost to the proposed method. The time complexity of the SK algorithm is an active field of research; classical algorithms, such as the \cite{orlin-flow, interior-point-methods}, find solutions to the unregularised OT problem in $\mathcal{O}\left(\eta^3\log(\eta)\right)$. \cite{cuturi2013sinkhorn} showed that approximate solutions $Q$ to the true minimum $\Tilde{Q}$, such that $|Q{-}\Tilde{Q}|{<}\tau$, can be found relatively fast by solving the corresponding entropic regularised problem in \eqref{eq:OT}. In particular, \cite{Altschuler2017} provided an upper bound for the number of the arithmetic operations needed for convergence of the entropy-regularised SK algorithm: $\mathcal{O}\left(\eta^3\log(\eta)/\tau^3 \right)$. Although without theoretical guarantees about the time complexity bound, \cite{Altschuler2017} also provided a version of the SK algorithm, named \textsc{GreenKhorn}, which empirically performs better than the implementation by \cite{cuturi2013sinkhorn}. \cite{Dvurechensky2018} recently improved on the results in \cite{Altschuler2017}  providing an algorithm with time complexity of $\mathcal{O}\left(\min\{\frac{\eta^9/4}{\tau}, \frac{\eta^2}{\tau^2}\} \right)$. In our method, we use the SK implementation by \cite{cuturi2013sinkhorn}, inducing a worst-case-scenario cost of $\mathcal{O}(T\left(\eta^3\log(\eta)/\tau^3 \right))$ for a dateset of dimension $T$, at each epoch.

 
\typeout{-------------- Other Experiments  --------------}
\section{Per-class accuracy in PascalVOC2012 and MS-COCO2014}\label{sec:tables}
This section presents additional tables for our results on the PascalVOC2012 dataset and  the MS-COCO2014 datasets.
In Table \ref{tab:Table1}, we present the accuracy for the weakly-supervised segmentation task on MS-COCO2014 {\em val} set for each class and compare our results with the methods which have the best accuracy so far, namely  MCTFormer \cite{xu2022multi}  and   ViT-PCM \cite{RossettiECCV-2022}. ViT-PCM  already had the total best accuracy w.r.t. MCTFormer of 3 points in mIoU\%, and we outperform ViT-PCM by about 1 point. 
In Table \ref{tab:COCO-Xclass}, we present the per-class mIoU\% accuracy of our PC2M on the unsupervised segmentation  task on MS-COCO2014 {\em val} set. Here we do not have available other methods to compare with. We can observe that for some classes such as 'baseball-bat', 'fork', 'spoon' and 'toaster', the mIoU accuracy is zero.

Table \ref{tab:WandUPVOC} reports the mIoU\% accuracy for the two weakly-supervised and unsupervised semantic segmentation tasks on PascalVOC2012 {\em val } set. We observe that there is no proportion between the two methods. In the unsupervised task, some classes like 'boat' are missed, and others classes like 'TV', 'aeroplane and 'bus' have an accuracy close to the one in the weakly-supervised task. Moreover, a common practice in WSSS is to  test the accuracy of the generated pseudo-segmentations by using them as masks to train a standard segmentation model and compare the latter with the ground-truth supervised version.  Interestingly, using our pseudo-masks to train DeepLabV3 \cite{chen2018encoder} leads to a less accurate model than our PC2M, respectively 74.2\% vs 75\% mIoU. Nevertheless, both results are still competitive for the supervised  DeepLabV3-JFT, which obtain 82.7\% mIoU on PascalVOC2012 \textit{val} set. See Table \ref{tab:WandUPVOC}.

\begin{table*}[h!]
\centering
\caption{Per-class performance comparison with the state-of-the-art WSSS methods  on the verification task with final-segmentation masks, in terms of mIoU\% on MS-COCO 2014 \textit{val} set. }\label{tab:Table1}
\resizebox{\textwidth}{!}{%
\begin{tabular}{l c  c  c  c | l  c  c  c  c } 
 \hline
 & Class & {\bf Ours (PC2M)} & MCT-Former\tiny{\textsc{CVPR'22} } & ViT-PCM\tiny{\textsc{ECCV'22} }  & &
 Class &  {\bf Ours (PC2M)} & MCT-Former\tiny{\textsc{CVPR'22} } & ViT-PCM\tiny{\textsc{ECCV'22} } \\
 & &    & \cite{xu2022multi}  &   \cite{RossettiECCV-2022}
 &  & & & \cite{xu2022multi} & \cite{RossettiECCV-2022} \\
         \hline
0. & background &84.5&82.4 & 81.9    
  & 41.&wine-glass&40.7&27.0& 38.2 
  \\
1. & person &59.5&62.6& 62.4 
   & 42.&cup&37.5&29.& 40.9 
   \\
2. & bicycle&50.7&47.4& 54.3  
 &43.&fork&8.1&13.9& 33.3
 \\
3. & car&48.7&47.2& 49.2
   &44.&knife&20.4&12.0& 31.0
   \\
4. & motorcycle&68.3&63.7& 70.3 
   &45.&spoon&8.7&6.6& 21.4
   \\
5. & airplane&70.1&64.7& 74.5
   &46.&bowl&42.8&22.4& 36.2
   \\
6. & bus&70.9&78.6& 76.0
   &47.&banana&65.3&63.2& 58.6
   \\
7. & train&58.9&64.5& 61.2
   &48.&apple&54.1&44.4& 52.1
   \\
8. &truck&48.5&44.8& 45.3
    &49.&sandwich&39.4&39.7& 57.1
    \\
9. & boat&35.0&42.3& 47.8
   &50.&orange&65.9&63.0& 55.8
   \\
10. & traffic-light&46.2&49.9& 22.2
   &51.&broccoli&30.5&51.2& 53.5
   \\
11. & fire-hydrant&77.7&73.2& 78.8
   &52.&carrot&28.9&40.0& 45.0
   \\
12. &stop-sign&78.5&76.6& 11.0
   &53.& hot-dog&65.5&53.0& 41.4
   \\
13.& parking-meter&71.8&64.4& 65.5
   &54.&pizza&79.3&62.2& 77.6
   \\
14. & bench&40.9&32.8& 42.6
   &55.&donut&75.2&55.7& 39.4
   \\
15. &bird&69.0&62.6& 67.0
    &56.&cake&34.6&47.9& 63.0
    \\
16. &cat&82.2&78.2& 20.4
    &57.&chair&24.6&22.8& 35.6
    \\
17. & dog &77.2&68.2& 71.7
    &58.&couch&41.5&35.0& 41.7
    \\
18. &horse&69.2&65.8& 68.6
    &59.&potted-plant&26.0&13.5& 37.9
    \\
19. &sheep&77.0&70.1& 67.2
    &60.&bed&53.5&48.6& 53.2 
    \\
20. & cow&76.4&68.3& 70.4
   &61.&dining-table&15.4&12.9& 29.4
   \\
21. & elephant&83.1&81.6&83.3 
    &62.&toilet&61.0&63.1& 67.3
    \\
22. &bear&85.0&80.1& 74.2
    &63.&tv &48.0&47.9& 38.7
    \\
23. &zebra&81.1&83.0 & 72.6
    &64.&laptop&58.0&49.5& 51.7
    \\
24. & giraffe&76.0&76.9& 67.3
    &65.&mouse&14.3&13.4& 13.9
    \\
25. &backpack&18.4&14.6& 24.3
    &66.&remote&41.2&41.9& 34.2
    \\
26.& umbrella&69.0&61.7& 67.7
     &67.&keyboard&60.4&49.8& 65.0
     \\
27.& handbag&7.2&4.5& 19.4
     &68.&cellphone&60.5&54.1& 56.8
     \\
28.& tie&21.3&25.2& 19.0
     &69.&microwave&28.0&38.0& 50.2
     \\
29.& suitcase&52.5&46.8 &47.6 
     &70.&oven&20.0&29.9& 35.8
     \\
30&. frisbee&24.0&43.8& 38.1
     &71.& toaster&4.6&0.0& 13.8
     \\
31. & skis&3.1&12.8& 20.3
     &72.&sink&30.4&28.0& 14.3
     \\
32. &snowboard&17.5&31.4&41.6 
     &73.&refrigerator&11.0&40.1& 44.9 
     \\
33. & sports-ball&8.0&9.2& 7.1
     &74.&book&38.1&32.2& 40.6
     \\
34& kite&60.2&26.3& 41.5
     &75.& clock&56.4& 43.2& 51.3
     \\
35. &baseball-bat&2.7&0.9 &2.3 
    &76.&vase&24.7&22.6& 25.0 
    \\
36. &baseball-glove&3.8&0.7& 5.0
    &77.& scissors&50.0&32.9& 48.1
    \\
37&skateboard&19.9&7.8& 10.3
    &78.&teddy-bear&67.9&61.9& 53.9
    \\
38& surfboard&34.9&46.5& 45.9
    &79.&hair-drier&20.9&0.0& 13.4
    \\
39. &tennis-racket&12.4&1.4& 16.1
    &80.&toothbrush&35.0&12.2& 33.1
    \\
\cline{6-10}
40. &bottle&39.0&31.1& 41.5
    & & mean & \textbf{45.92} & 42.0  & 45.0\\
    \hline
\end{tabular}
}\label{tab:coco_classes}
\end{table*}

\begin{table*}[h]
  \centering
  \caption{ Per class performance of our method on the task of predicting Unsupervised pseudo-labels and pseudo-masks, in terms of mIoU\%  on MS-COCO2014 {\em val} set. }
  \resizebox{\textwidth}{!}{%
 
     \begin{tabular}{l|cccc ccccccc}
     \hline \hline
    Classes & background & person & bicycle & car   & {motorcycle} & {airplane} & {bus} & {train} & {track} & {boat} & {traffic light} \\
    \rowcolor{gray!14}Accuracy &  69.0& 4.4 &35.8 & 1.1 & 55.4 & 32.8  & 51.2  & 23.1  & 25.1  & 17.5  & 39.6  \\
    \hline\hline
    Classes & fire-hydrant & stop-sign & parking-meter & bench & {bird} & {cat} & {dog} & \multicolumn{1}{c}{horse} & {ship} & {cow} & {elephant} \\
    \rowcolor{gray!15}Accuracy & 52.7 &  38.8 & 0.0 & 25.4 & 23.7  & 36.6  & 24.8  & 52.1  & 36.5  & 51.9  & 73.9 \\
    \hline \hline
    Classes & bear  & zebra & giraffe & backpack & {umbrella} & {handbag} & {tie} & {suitcase} & {frisbee} & {skis} & {snowboard} \\
   \rowcolor{gray!15} Accuracy & 3.3 & 80.6 & 76.1 & 0.9 & 50.8  & 1.1   & 11.1  & 1.1   & 1.1   & 0.7  & 3.8 \\
    \hline
    \hline
    Classes & sports-ball & kite  & baseball-bat & baseball-glove & {skateboard} & {surfboard} & {tennis-racket} & {bottle} & {wine-glass} & {cup} & {fork} \\
    \rowcolor{gray!15} Accuracy &0.0 & 3.6 & 0.1 & 0.7 & 8.4 & 1.6   & 2.9   & 3.4   & 0.0   & 0.0  & 0.0  \\
     \hline\hline
    Classes & knife  & spoon & ball  & banana & {apple} & {sandwich} & {orange} & {broccoli} & carrot & {hot-dog} & {pizza}  \\
    \rowcolor{gray!15}Accuracy &0.0 &0.0 & 14.5 & 58.8 & 0.0     & 36.4  & 24.0    & 16.1  & 0.0  & 0.3   & 66.1 \\
     \hline\hline
    Classes &{donut} & cake  & chair & coach & potted-plant & {bed} & {dining-table} & {toilet} & {tv} & {laptop} &   {mouse} \\
    \rowcolor{gray!15}Accuracy & 23.7 & 11.1 & 0.3 & 9.2 & 0.1  & 10.0    & 2.1   & 56.8  & 15.4  & 25.2  & 0.0\\
     \hline \hline
    Classes & remote & keyboard & cell  & microwave & {oven} & {toaster} & {sink} & {refrigerator} & {book} & {clock} & {vase}\\
    \rowcolor{gray!15}Accuracy & 0.1 & 0.0 & 17.8 & 14.6 & 4.5   & 0.0     & 12.8  & 0.1   & 7.8   & 49.1  & 22.4  \\
    \hline\hline
    Classes &scissors & teddy-bear & hair-dryer & tooth-brush &       &       &       &       &       &       & mean \\
    \rowcolor{gray!15}Accuracy & 6.2 & 56.7 & 0.0 & 0.3 &       &       &       &       &       &       &  19.5 \\
    \hline\hline
    \end{tabular}%

  \label{tab:COCO-Xclass}%
  }
\end{table*}%

\begin{table*}[h!]
\centering
\caption{Per-class performance in terms of mIoU\% for both weakly-supervised and unsupervised segmentation tasks on PascalVOC2012 {\em val} dataset. }
\label{tab:WandUPVOC}
\resizebox{\textwidth}{!}{%
\begin{tabular}{c | c  c  c  c  c  c  c  c  c  c  c  c  c  c  c  c  c  c  c  c c| c} 
 \hline

 Task & bkg & aero & bike & bird & boat & btl & bus & car & cat & chair & cow & table & dog & horse & mbk & person & plant & sheep & sofa & train & tv & mean \\ 
 \hline
 WSSS & 92.8 & 87.4 & 49.0 & 90.7 & 73.9 & 77.2 & 87.4 & 82.0 & 87.3 & 40.7 & 86.5 & 33.3 & 89.2 & 88.7 & 82.5 & 77.2 & 66.6 & 91.7 & 49.4 & 78.5 & 62.5 &75.0\\
  WSSS+DeepLabV3\cite{chen2018encoder} & 92.7 & 86.8 & 45.8 & 82.6 & 70.7 & 75.4 & 92.4 & 83.8 & 90.7 & 42.2 & 86.2 & 41.2 & 89.2 & 86.2 & 78.7 & 77.4 & 57.1 & 84.7 & 45.0 & 82.0 & 66.5 &74.2\\
 USS  &  86.6 &73.1 &41.5 &79.1 &0.0 &40.3 &70.0 &66.8 &69.3 &19.6 &0.06 &17.1 &40.2 &58.7 &63.1 &17.8 &0.03 &43.9 &0.0 &42.7 &50.6 &43.6 \\
 \hline
  \hline
\end{tabular}
}
\end{table*}

\typeout{-------------Qualitative comparisons with other methods--------------}
\section{Qualitative results and qualitative comparisons with other methods}\label{sec:Unsup_method}
This section compares our results with other weakly-supervised and unsupervised semantic segmentation methods on PascalVOC2012 {\em val} set and on MS-COCO2014 {\em val} set and shows our qualitative results.
\noindent
\subsection{Qualitative comparisons on PascalVOC2012}
\begin{figure*}[h]
\centering
 \includegraphics[width=0.99\linewidth]{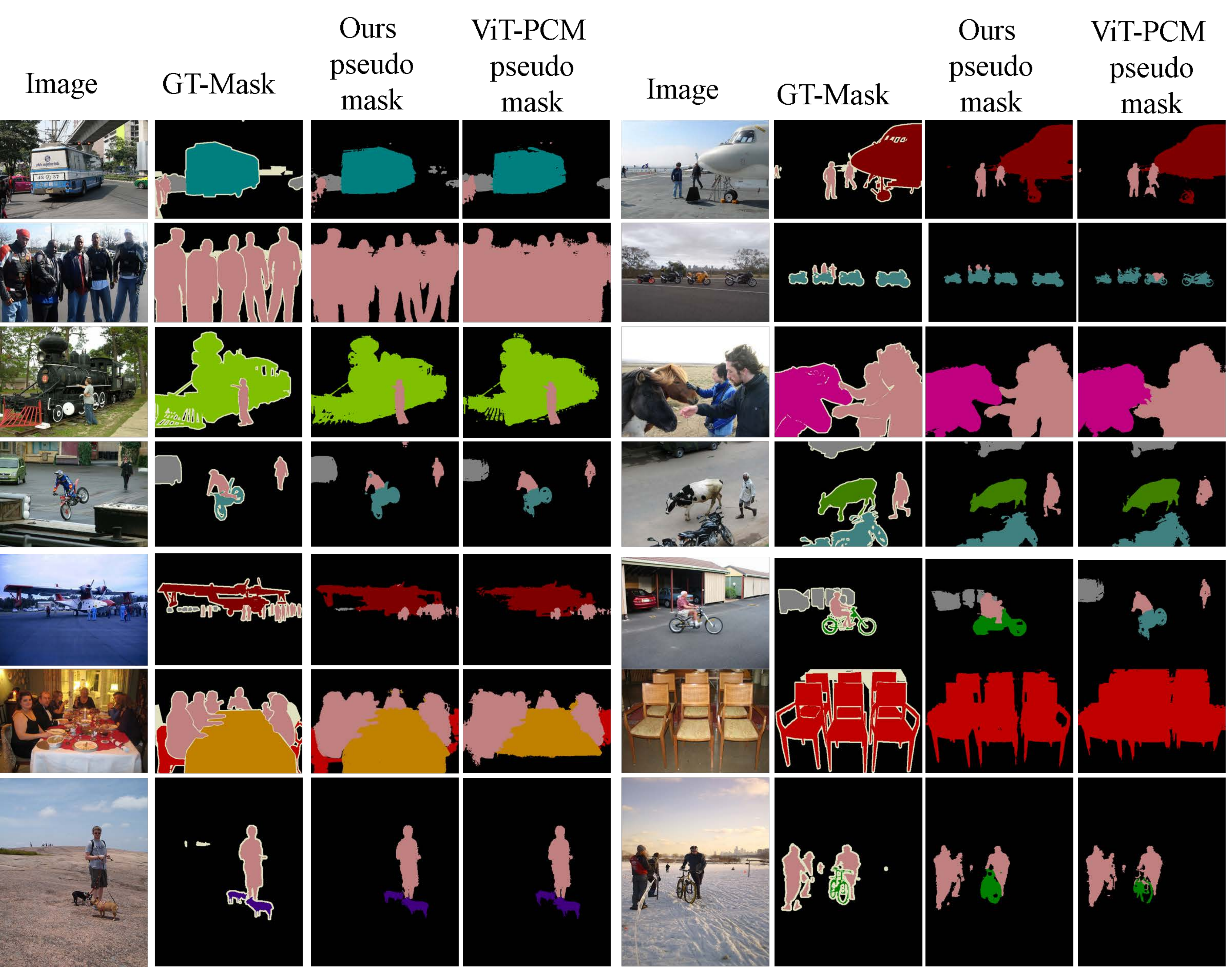}
\caption{Comparison with state-state-of-the-art results of ViT-PCM \cite{RossettiECCV-2022}  for weakly-supervised Semantic Segmentation on PascalVOC val set. From left to right: Image, colour-coded ground truth mask, with the PascalVOC palette, ours, ViT-PCM \cite{RossettiECCV-2022}. }
\label{fig:CompVItPCM}
\end{figure*}

\begin{figure*}[h]
\centering
 \includegraphics[width=0.9\linewidth]{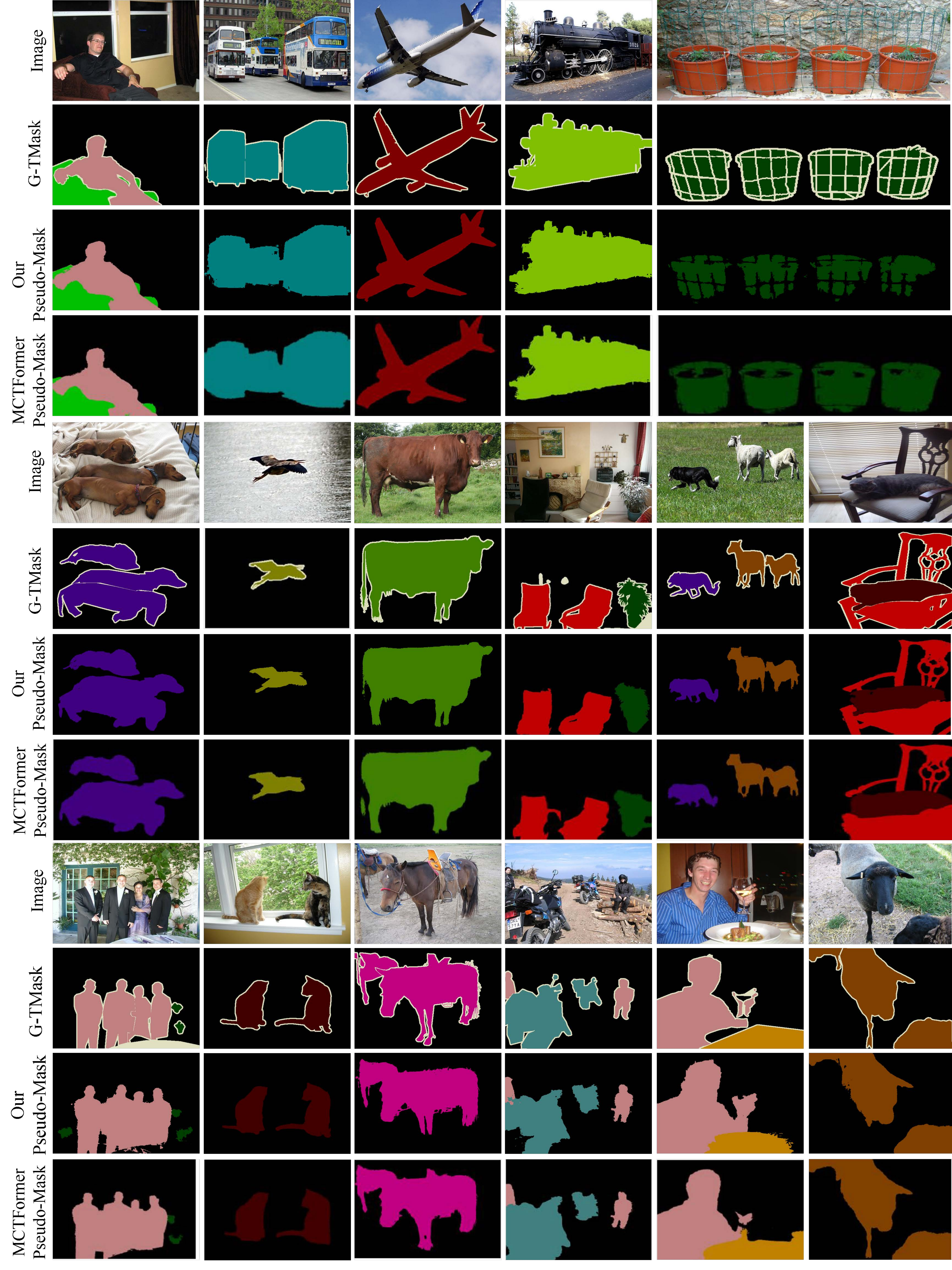}
\caption{Comparison with MCT Former \cite{xu2022multi} on PascalVOC val dataset. The pseudo-masks of MCTFormer \cite{xu2022multi} are taken from the paper's supplementary materials. Ivory parts indicate no class. }
\label{fig:Comp_MCT_former}
\end{figure*}

\noindent
{\bf Weakly-supervised task.}
To assess our qualitative results, we compare the pseudo-masks obtained with our method with ViT-PCM \cite{RossettiECCV-2022} and MCTFormer \cite{xu2022multi}. ViT-PCM and MCTFormer currently get the best accuracy on weakly supervised segmentation. Results are illustrated in Figure \ref{fig:CompVItPCM}  and Figure \ref{fig:Comp_MCT_former}. 
Concerning MCTFormer \cite{xu2022multi}, note that we have taken the images to make the  comparison  from the paper \cite{xu2022multi}; therefore, we have not chosen our best results. Still, in several images, it is clear that our method obtains more refined pseudo-masks than MCTFormer.

\begin{figure*}[h]
\centering
 \includegraphics[width=0.85\linewidth]{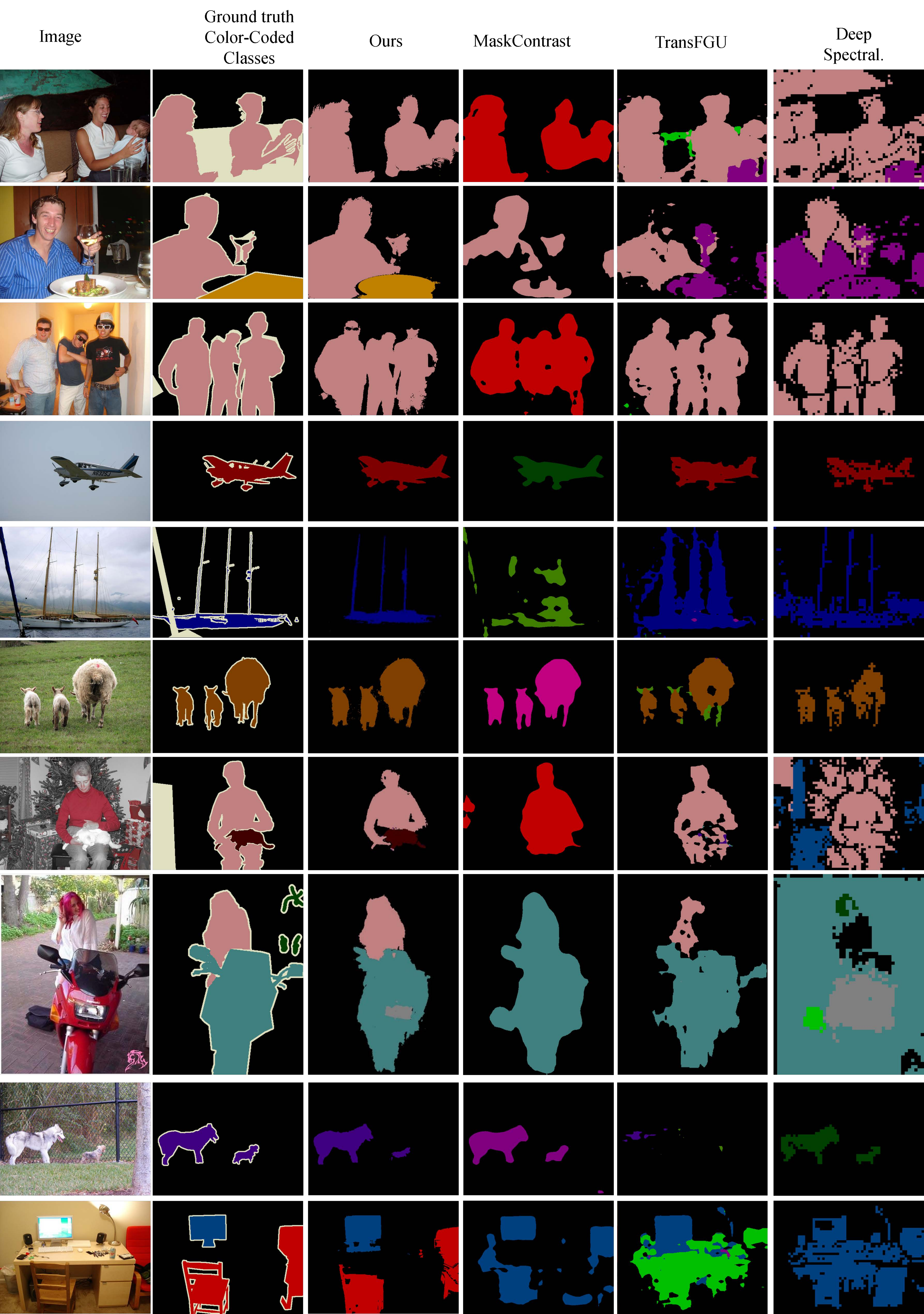}
\caption{Comparison with state-state-of-the-art results of Unsupervised Semantic Segmentation on PascalVOC {\em val} set. From left to right: Image, colour-coded ground truth mask, with the PascalVOC palette, ours, MaskContrast \cite{van2021unsupervised}, TransFGU\cite{yin2022transfgu}, Spectral Segmentation \cite{melas2022deep}.}
\label{fig:CompUnsupervised1}
\end{figure*}

\begin{figure*}[h]
\centering
 \includegraphics[width=0.95\linewidth]{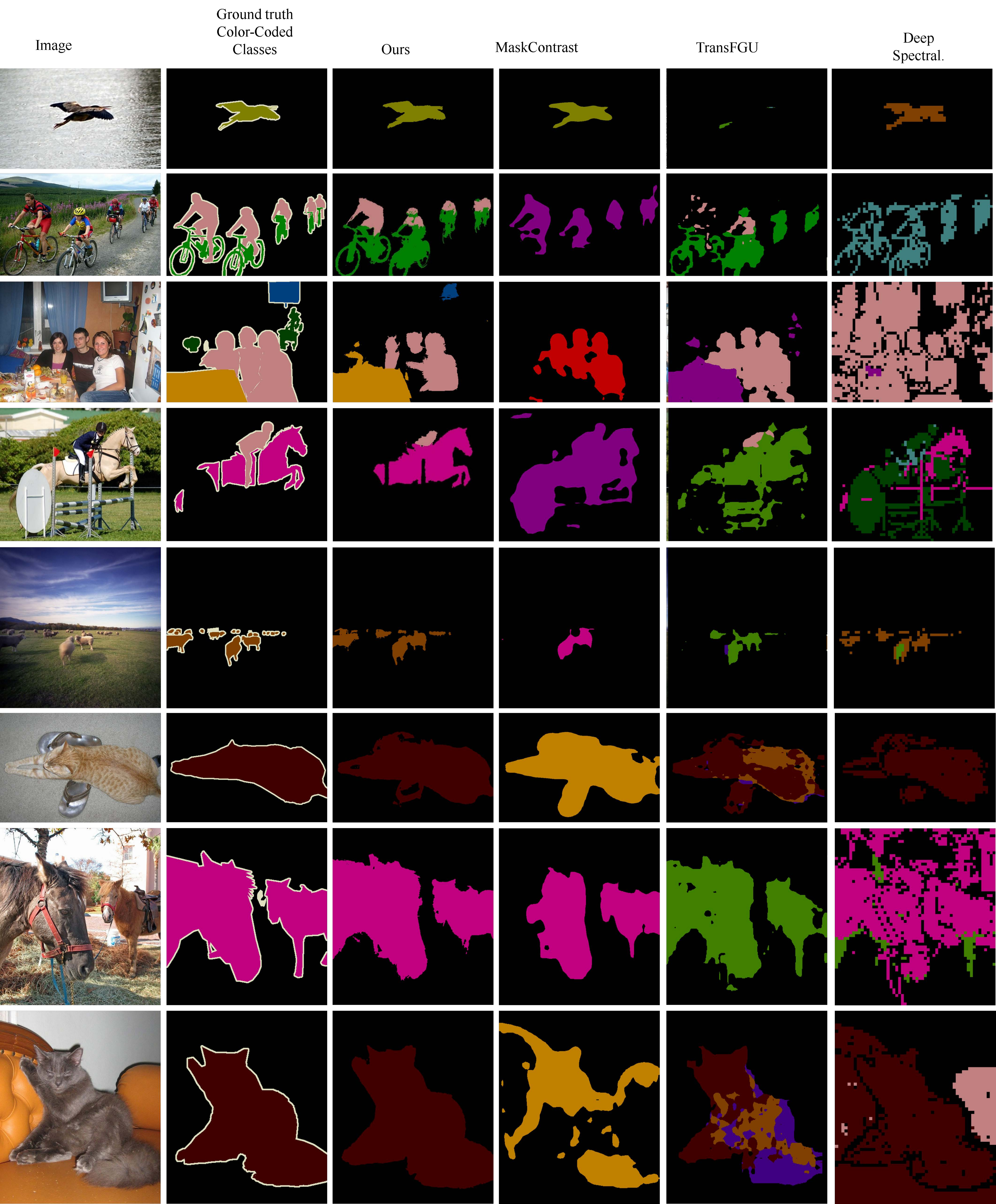}
\caption{Comparison with state-state-of-the-art results of Unsupervised Semantic Segmentation on PascalVOC. From left to right: Image, colour-coded ground truth mask, with the PascalVOC palette, ours, MaskContrast \cite{van2021unsupervised}, TransFGU\cite{yin2022transfgu}, and Spectral Segmentation \cite{melas2022deep}.}
\label{fig:CompUnsupervised2}
\end{figure*}

\begin{figure*}[h]
\centering
 \includegraphics[width=0.95\linewidth]{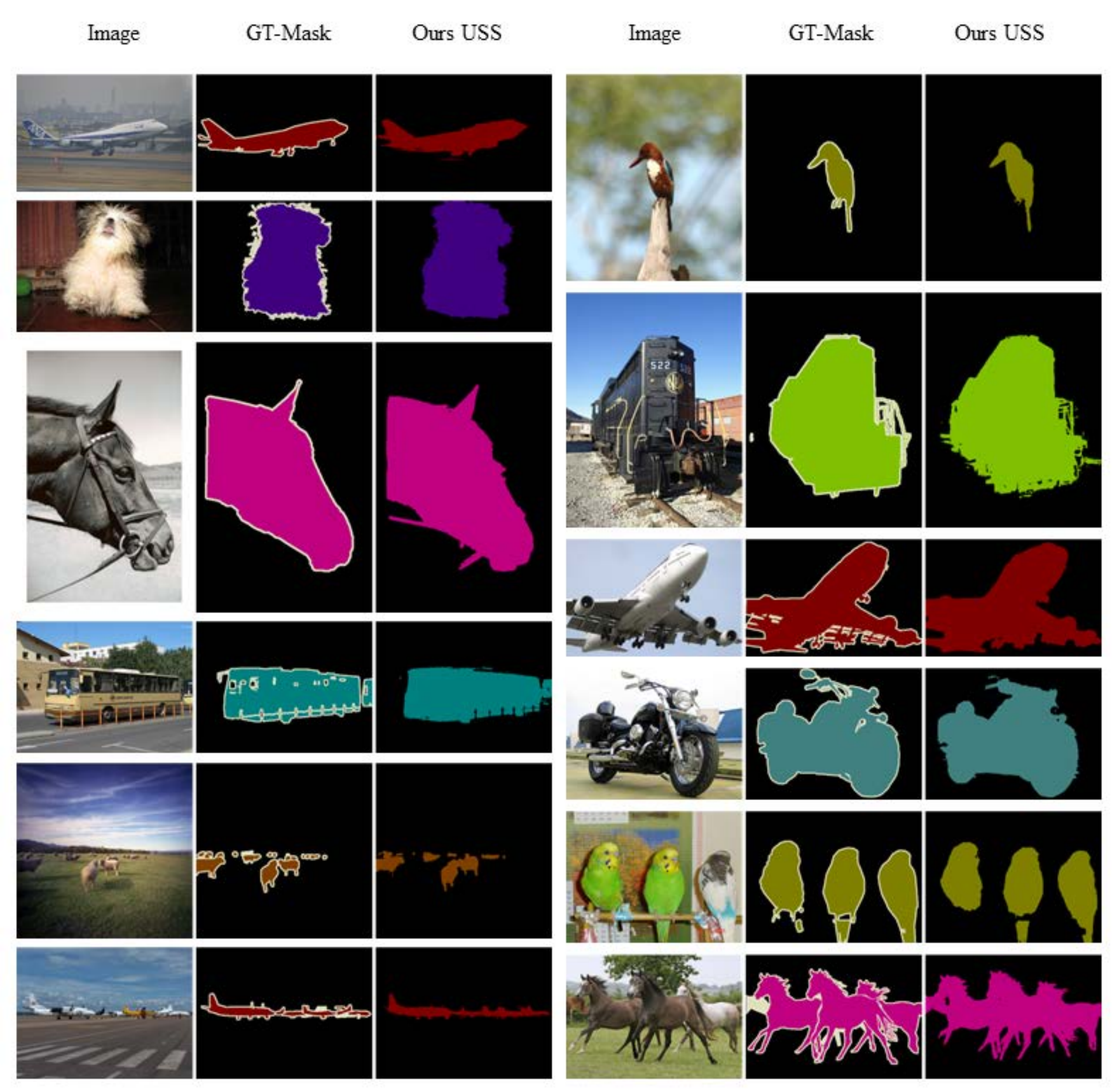}
\caption{Our best mIoU\% accuracy results for Unsupervised Semantic Segmentation on PascalVOC {\em val} set}
\label{fig:ourUnsupPV}
\end{figure*}

\noindent 
{\bf Unsupervised task.}
Here, we compare our results with three other approaches to unsupervised semantic segmentation.
To produce the pseudo-mask and pseudo-labels, we have operated the publicly available software of  MaskContrast \cite{van2021unsupervised}, TransFGU\cite{yin2022transfgu}, and Deep Spectral Segmentation \cite{melas2022deep}. 
First of all, we have verified that the declared results for each competitor were satisfied. In particular, for both MaskContrast and Deep Spectral Segmentation \cite{melas2022deep}, we have used the PascalVOC2012 class colour palette to make it comparable with the other methods. The mapping to the PascalVOC2012 category set is obtained via the Hungarian algorithm to maximize the total mIoU\%. Among our images returning correct classes, we have chosen those for which at least one of our competitors' methods had a good shape or correct classes.   
See Figures \ref{fig:CompUnsupervised1} and \ref{fig:CompUnsupervised2}. Figure \ref{fig:ourUnsupPV} shows unsupervised masks on PascalVOC2012 {\em val} set.

\subsection{Qualitative results  on MS-COCO2014 {\em val} set} 
\begin{figure*}[h]
\centering
 \includegraphics[width=0.85\linewidth]{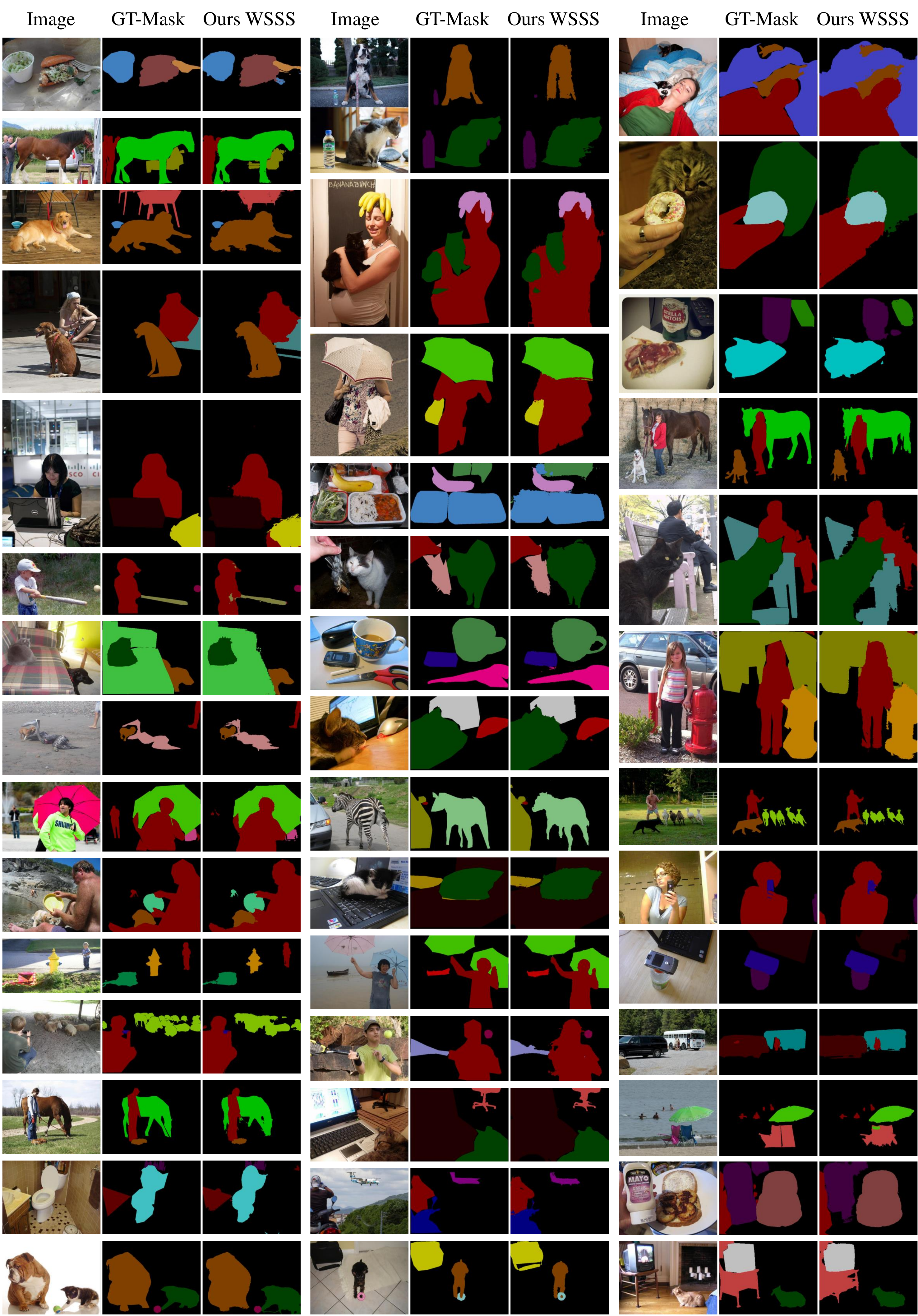}
\caption{Our qualitative results for the WSSS task on COCO2014 {\em val} set.}
\label{fig:CocoWSSS}
\end{figure*}

\begin{figure*}[h]
\centering
 \includegraphics[width=0.85\linewidth]{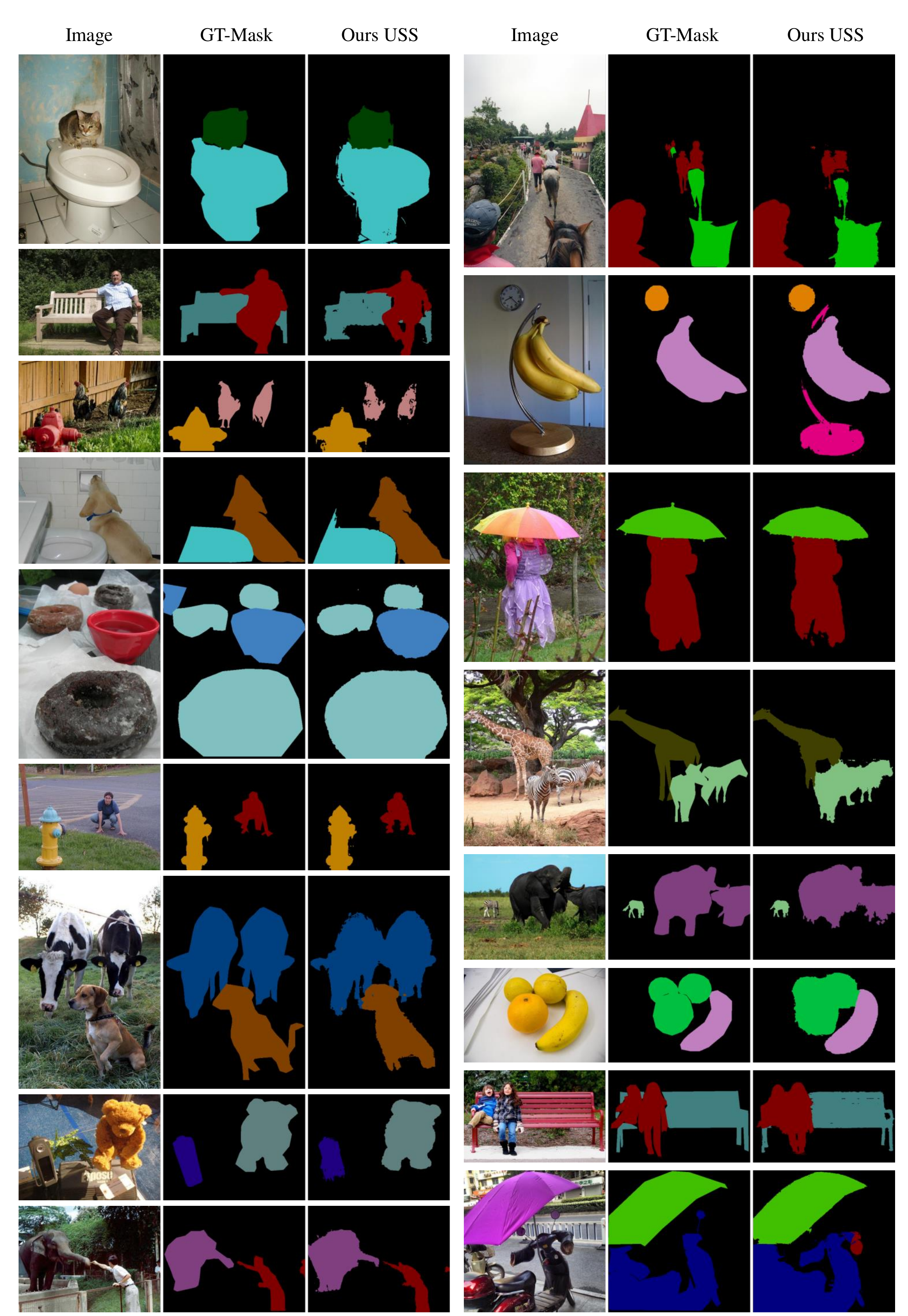}
\caption{Our qualitative results for unsupervised semantic segmentation on COCO2014 {\em val} set.}
\label{fig:CocoUSS}
\end{figure*}

\begin{figure*}[h]
\centering
 \includegraphics[width=0.95\linewidth]{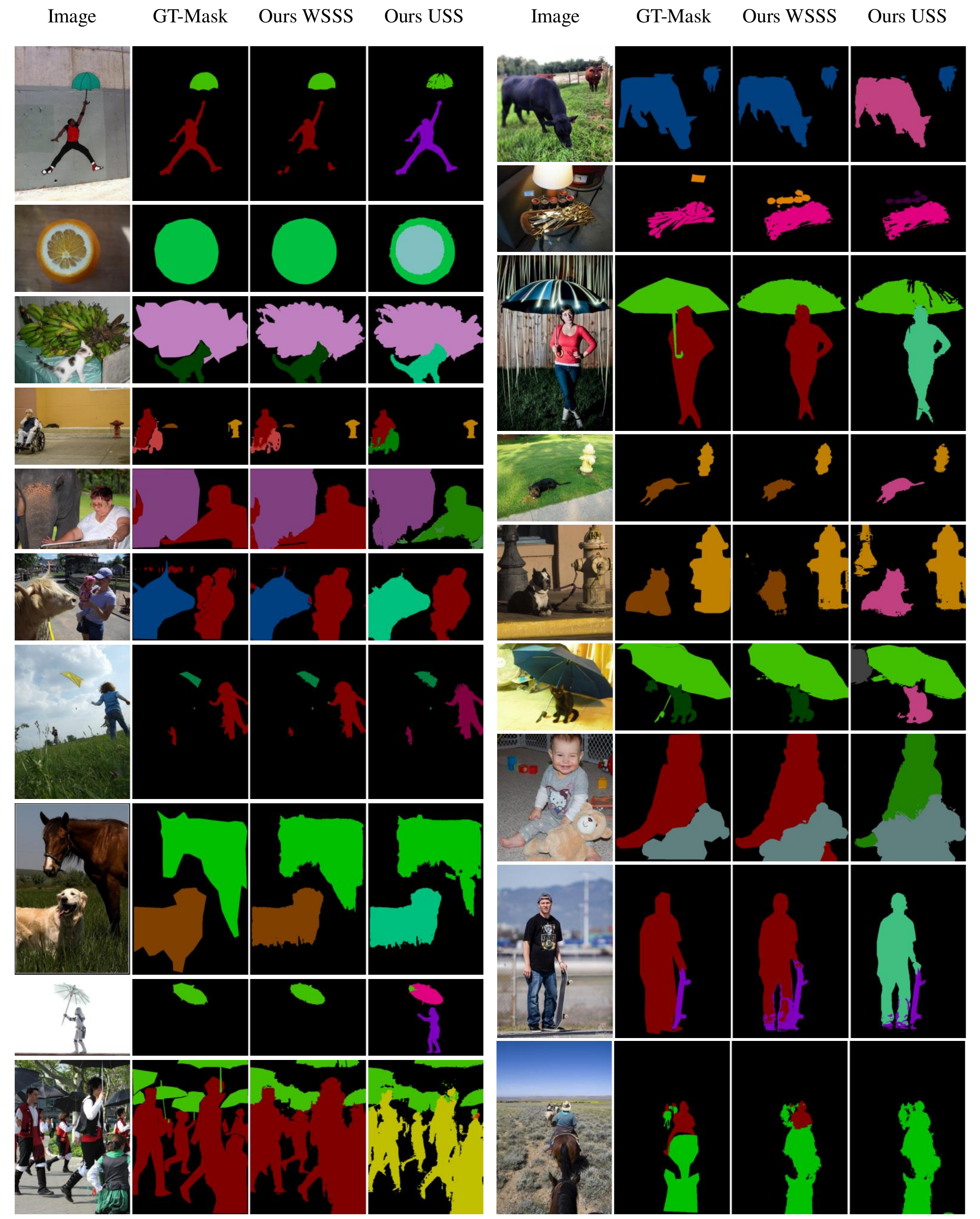}
\caption{The Figure shows qualitative comparisons of our weakly-supervised and unsupervised semantic segmentation results on COCO 2014 \textit{val}. In this picture, we intentionally show classification errors of our USS method. While in both methods, segmentation masks are usually coherent with the ground truth, objects are usually miss-classified in an unsupervised setting, as the "Ours USS" column shows, pointing to possible improvements in the image label clustering phase. In fact, introducing ground truth image-level labels solve the classification errors, as shown in the "Ours WSSS" column.}
\label{fig:uss_mismatch}
\end{figure*}

In Figure \ref{fig:CocoUSS}, we show our best mIoU\% accuracy  results for the WSSS task, while in Figure \ref{fig:CocoWSSS} we report our results on the USS task, on MS-COCO2014 {\em val} set. Finally, in Figure \ref{fig:uss_mismatch}
we show qualitative comparisons of our weakly-supervised and unsupervised semantic segmentation results on MS-COCO2014 \textit{val}. We intentionally show results in which segmentation masks are coherent for both methods, but objects are sometimes misclassified in the unsupervised setting, pointing to possible improvements in the image label clustering phase. Indeed, we observe low values for the generated pseudo multi-labels accuracy w.r.t. the ground-truth multi-labels, namely 42\% macro F1-score and 37\% micro F1-score on the MS-COCO2014 \textit{train} set. Similarly, on the PascalVOC2012 \textit{trainaug} set, we observe  53\% macro F1-score and 51\% micro F1-score.

\typeout{-------------Metric--------------}
\section{Considerations on the USS metrics}\label{sec:Unsup_method2}
Currently used metrics for USS mix-up labels with shape accuracy.
We would like to measure the shape accuracy for unsupervised semantic segmentation independently from the multi-label one. 
For each image $X{\in}{\mathbb R}^{h{\times} w {\times} 3}$, we consider the ground truth masks and the network predicted masks dropping the relative class ids. Thus we consider two sets of binary masks of cardinality $|C|$. We use $IoU_w$ as the cost weight of the edges of a bipartite graph:
\begin{equation}
    IoU_{w} = \frac{|M_{GT}^i\cap M_P^j|}{|M_{GT}^i\cup M_P^j|} \mbox{ with } i,j=1,\dots,|C|.
\end{equation}
We solve the max cost associated with the graph, which does not imply the max match of the vertices. We take the average of all costs $IoU_w$ for each image in the validation set, obtaining $mIoU_{shape}$.
This metric clearly does not account for the semantics given by the multi-labels, but it singles out the geometric content of the image, which is a crucial aspect of USS. 
For our method, the $mIoU_{shape}$ over all images is 65.9\% on PascalVOC2012 and 38.2\% on MS-COCO2014 \textit{val} sets. 


\end{appendices}

\end{document}